\documentclass[journal]{IEEEtran}
\usepackage[utf8]{inputenc}
\usepackage[caption=false,font=footnotesize]{subfig}
\usepackage[nopdftrailerid=true]{pdfprivacy}
\usepackage{textgreek}
\usepackage{amssymb}
\usepackage{tikz}
\usepackage{tikz-3dplot}
\usepackage{bm}
\usepackage{dsfont}
\usepackage{datatool}
\usepackage{tabularx}
\usepackage{booktabs}
\usepackage{adjustbox}
\usepackage{algorithm}
\usepackage{algorithmic}
\usetikzlibrary{positioning}
\usetikzlibrary{arrows.meta}
\usepackage{hyperref}

\newcommand\figscale{0.15}
\begin{document}
\tdplotsetmaincoords{100}{-70}
\begin{filecontents*}{python/tmp/keys-values.csv}
key,value
uci-epilepsy-supervised-accuracy,82.10
mnist-supervised-accuracy,92.07
fashionmnist-supervised-accuracy,81.45
\end{filecontents*}
\DTLloaddb{keys-values}{python/tmp/keys-values.csv}

\title{Sparsely Activated Networks}
\author{Paschalis~Bizopoulos, and~Dimitrios~Koutsouris
\thanks{P. Bizopoulos and D. Koutsouris are with Biomedical Engineering Laboratory, School of Electrical and Computer Engineering, National Technical University of Athens, Greece e-mail: pbizop@gmail.com}}

\maketitle

\begin{abstract}
	Previous literature on unsupervised learning focused on designing structural priors with the aim of learning meaningful features.
	However, this was done without considering the description length of the learned representations which is a direct and unbiased measure of the model complexity.
	In this paper, first we introduce the $\varphi$ metric that evaluates unsupervised models based on their reconstruction accuracy and the degree of compression of their internal representations.
	We then present and define two activation functions (Identity, ReLU) as base of reference and three sparse activation functions (top-k absolutes, Extrema-Pool indices, Extrema) as candidate structures that minimize the previously defined $\varphi$.
	We lastly present Sparsely Activated Networks (SANs) that consist of kernels with shared weights that, during encoding, are convolved with the input and then passed through a sparse activation function.
	During decoding, the same weights are convolved with the sparse activation map and subsequently the partial reconstructions from each weight are summed to reconstruct the input.
	We compare SANs using the five previously defined activation functions on a variety of datasets (Physionet, UCI-epilepsy, MNIST, FMNIST) and show that models that are selected using $\varphi$ have small description representation length and consist of interpretable kernels.
\end{abstract}

\begin{IEEEkeywords}
	artificial neural networks, autoencoders, compression, sparsity
\end{IEEEkeywords}

\section{Introduction}
Deep Neural Networks (DNNs)~\cite{lecun2015deep} use multiple stacked layers containing weights and activation functions, that transform the input to intermediate representations during the feed-forward pass.
Using backpropagation~\cite{rumelhart1986learning} the gradient of each weight w.r.t.\ the error of the output is efficiently calculated and passed to an optimization function such as Stochastic Gradient Descent or Adam~\cite{kingma2014adam} which updates the weights making the output of the network converge to the desired output.
DNNs were successful in utilizing big data and powerful parallel processing units and achieved state-of-the-art performance in problems such as image~\cite{krizhevsky2012imagenet} and speech recognition~\cite{graves2013speech}.
However, these breakthroughs have come at the expense of increased description length of the learned representations, which in sparsely represented DNNs is proportional with the number of:
\begin{itemize}
	\item weights of the model and
	\item non-zero activations.
\end{itemize}

The use of large number of weights as a design choice in architectures such as Inception~\cite{szegedy2016rethinking}, VGGnet~\cite{simonyan2014very} and ResNet~\cite{he2016deep} (usually by increasing the depth) was followed by research that signified the weight redundancy of DNNs.
It was demonstrated that DNNs easily fit random labeling of the data~\cite{zhang2016understanding} and that in any DNN there exists a subnetwork that can solve the given problem with the same accuracy with the original one~\cite{frankle2018lottery}.

Moreover, DNNs with large number of weights have higher storage requirements and they are slower during inference.
Previous literature addressing this problem has focused on weight pruning from trained DNNs~\cite{aghasi2017net} and weight pruning during training~\cite{lin2017runtime}.
Pruning minimizes the model capacity for use in environments with low computational capabilities, or low inference time requirements and helps reducing co-adaptation between neurons, a problem which was also addressed by the use Dropout~\cite{srivastava2014dropout}.
Pruning strategies however, only take in consideration the number of weights of the model.

The other element that affects the description length of the representation of DNNs is the number of non-zero activations in the intermediate representations which is related with the concept of activity sparseness.
In neural networks sparseness can be applied on the connections between neurons, or in the activation maps~\cite{laughlin2003communication}.
Although sparseness in the activation maps is usually enforced in the loss function by adding a $L_{1, 2}$ regularization or Kullback-Leibler divergence term~\cite{kingma2013auto}, we could also achieve sparsity in the activation maps with the use of an appropriate activation function.

\begin{figure*}[!t]
	\centering
	\subfloat{\includegraphics[width=0.2\textwidth]{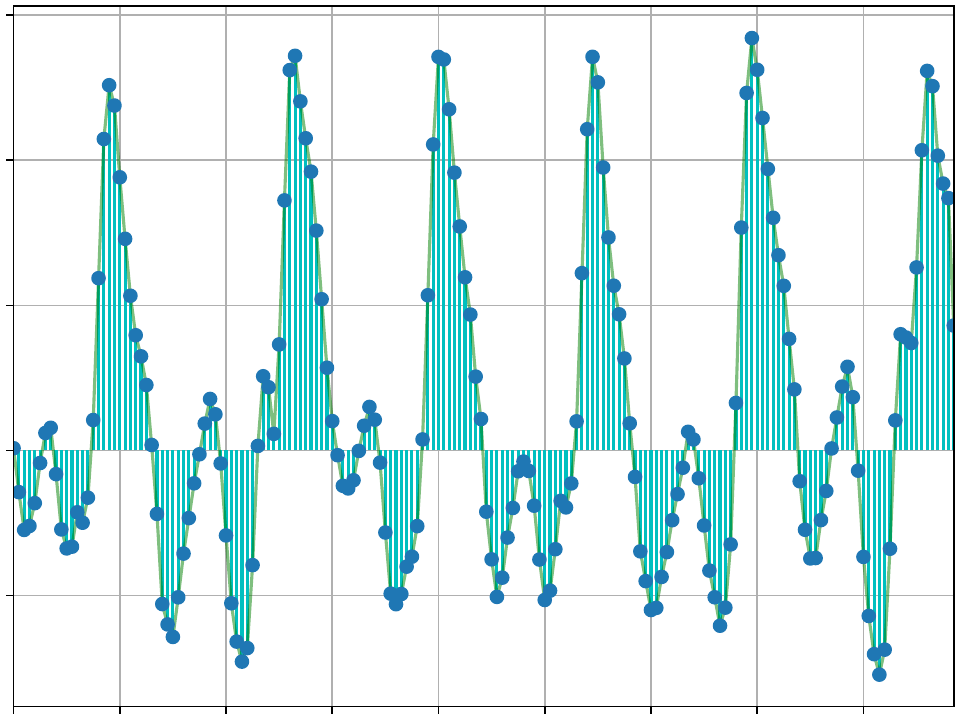}}
	\subfloat{\includegraphics[width=0.2\textwidth]{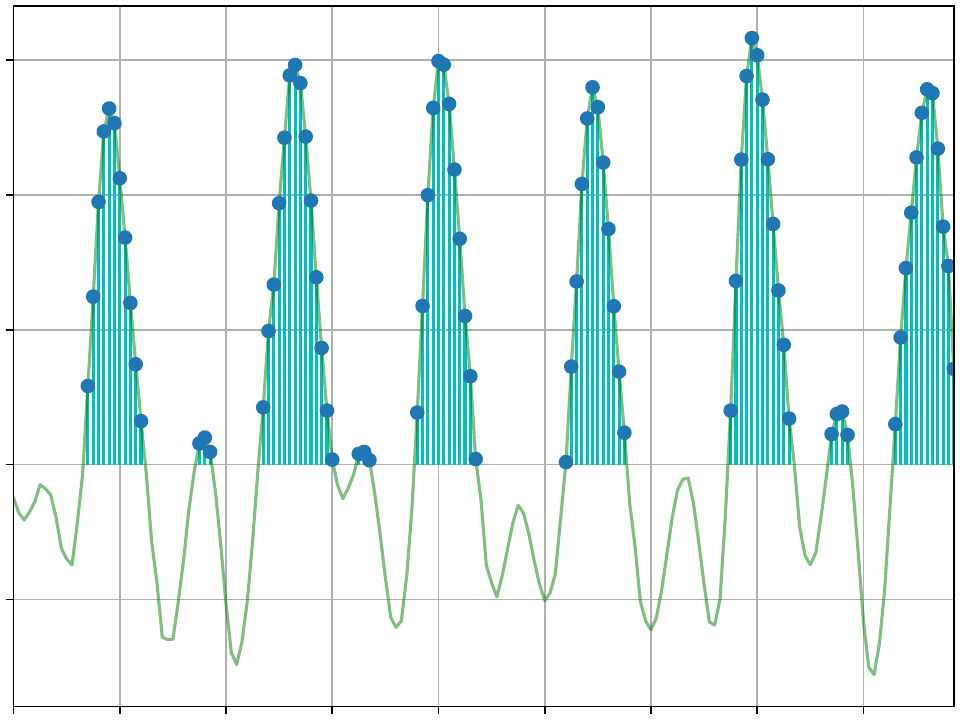}}
	\subfloat{\includegraphics[width=0.2\textwidth]{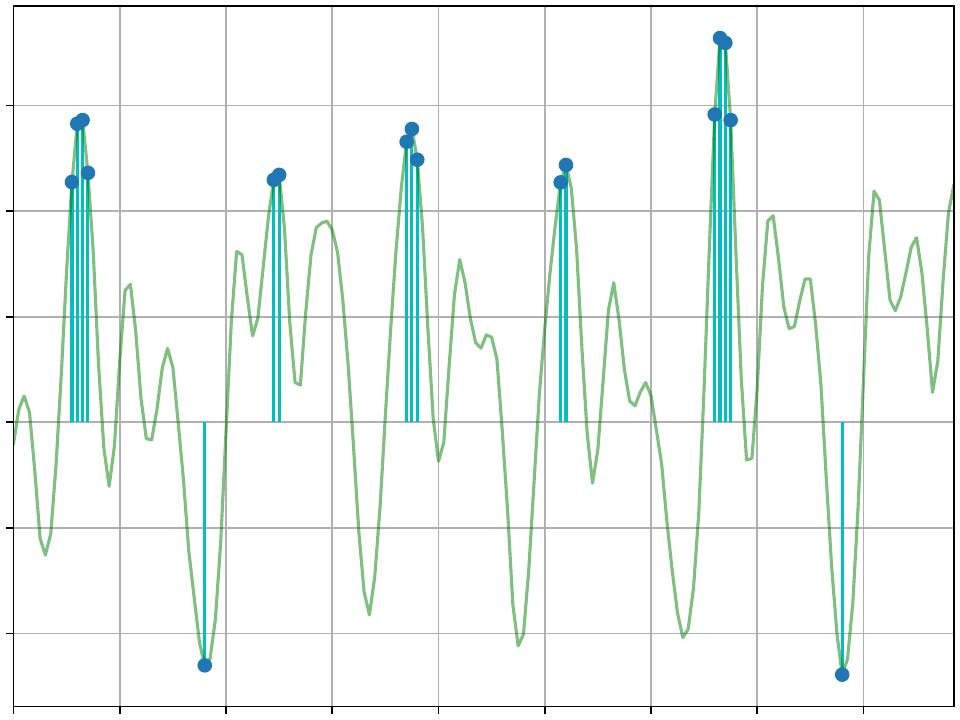}}
	\subfloat{\includegraphics[width=0.2\textwidth]{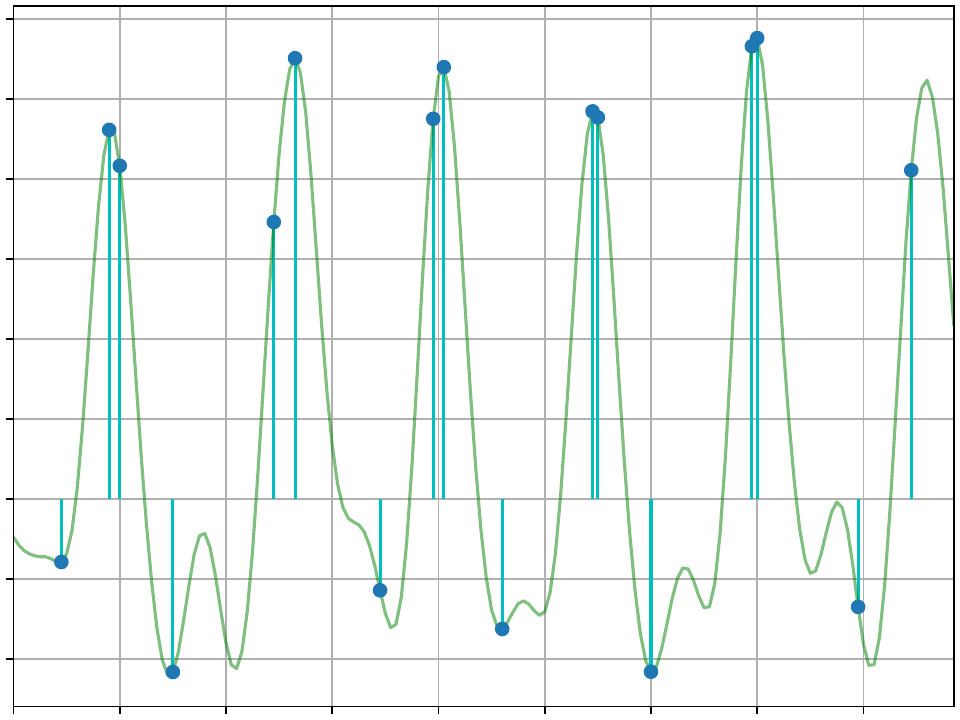}}
	\subfloat{\includegraphics[width=0.2\textwidth]{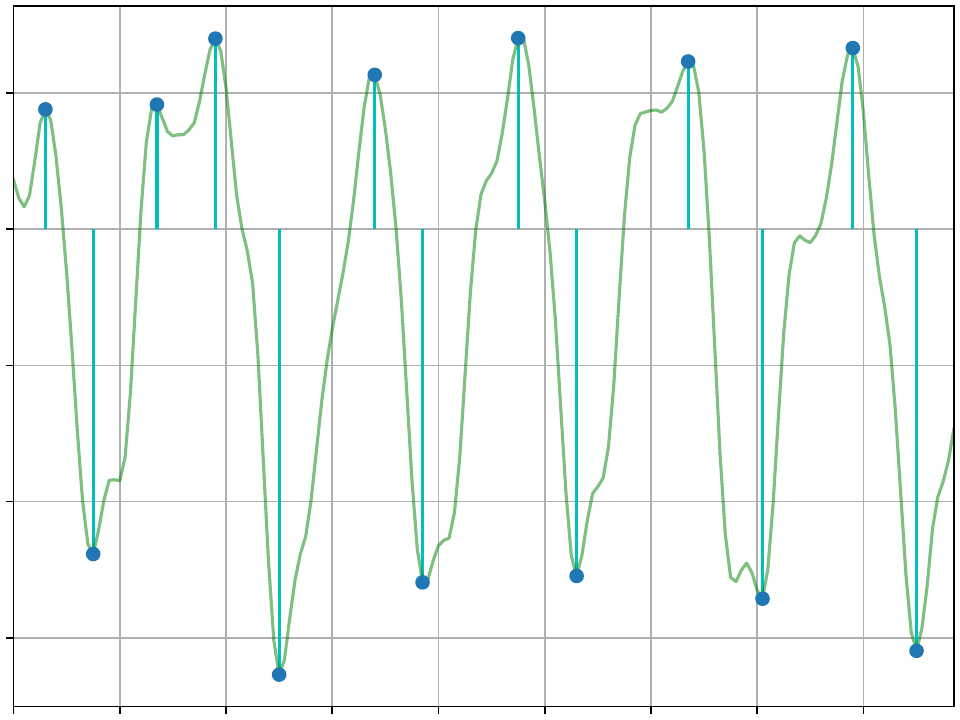}}
	\\
	\setcounter{subfigure}{0}
	\subfloat[Identity]{\includegraphics[width=0.2\textwidth]{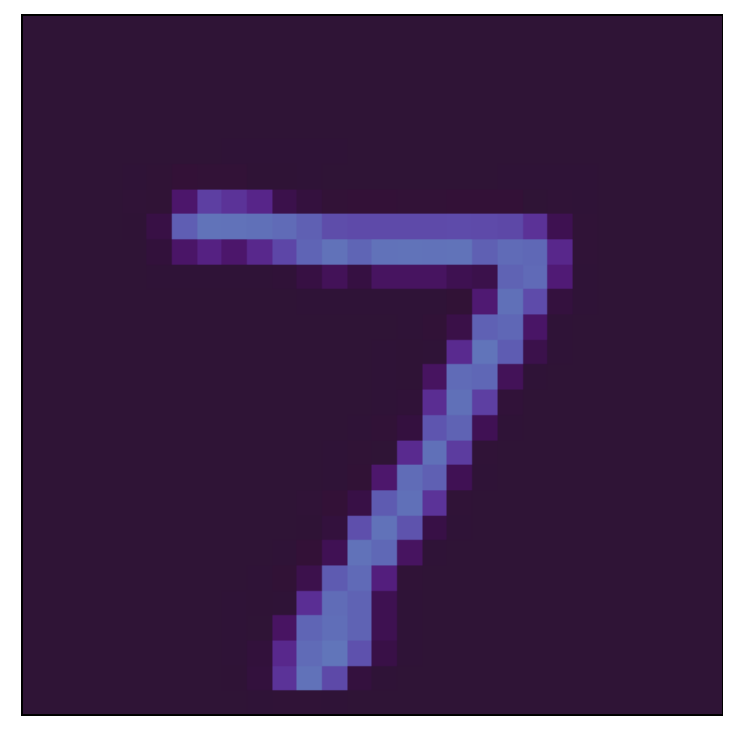}\label{subfig:identity}}
	\subfloat[ReLU]{\includegraphics[width=0.2\textwidth]{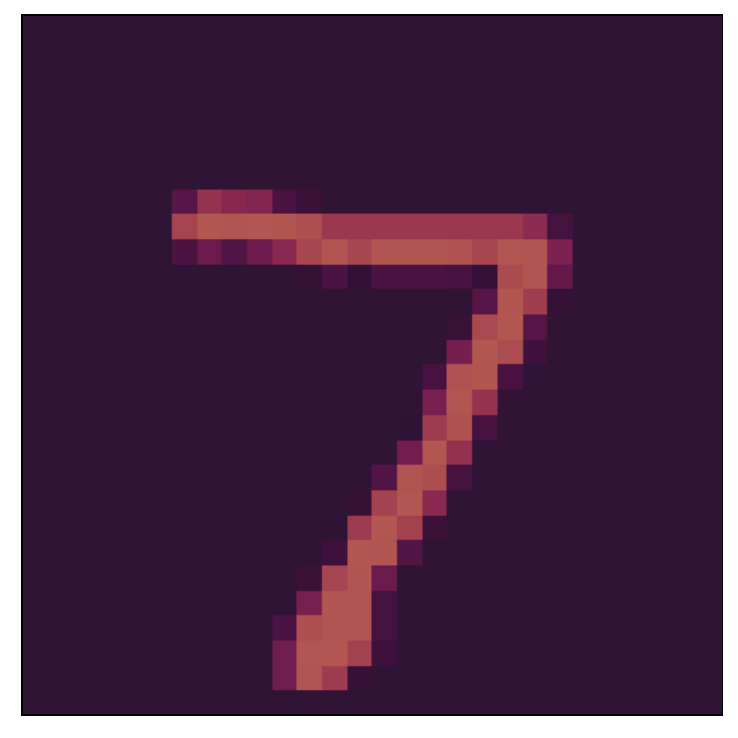}\label{subfig:relu}}
	\subfloat[top-k absolutes]{\includegraphics[width=0.2\textwidth]{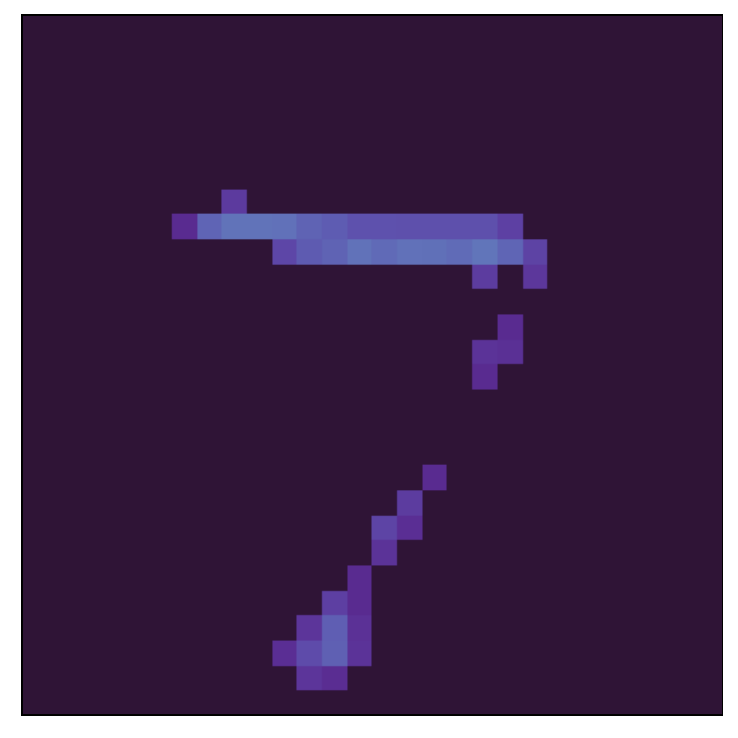}\label{subfig:topkabsolutes}}
	\subfloat[Extrema-Pool indices]{\includegraphics[width=0.2\textwidth]{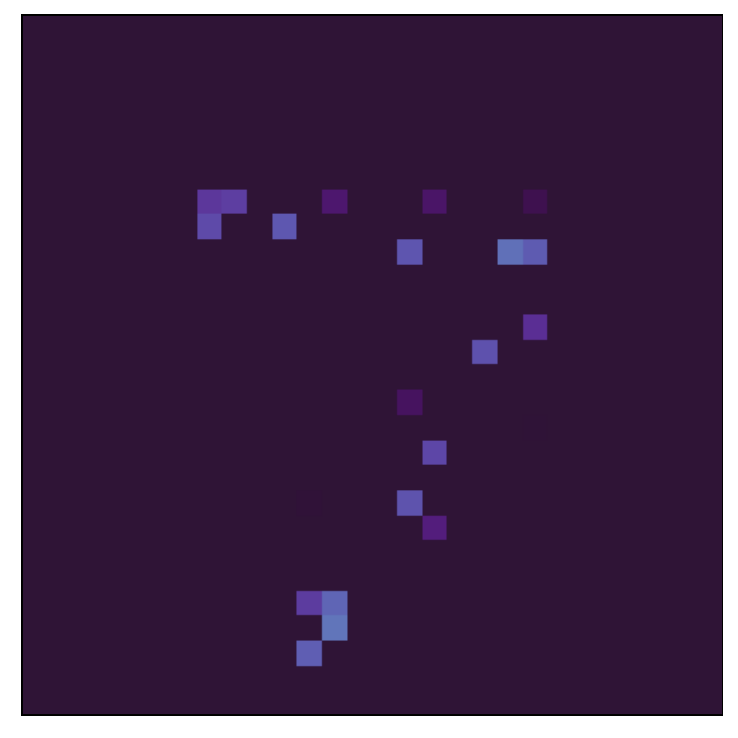}\label{subfig:extremapoolindices}}
	\subfloat[Extrema]{\includegraphics[width=0.2\textwidth]{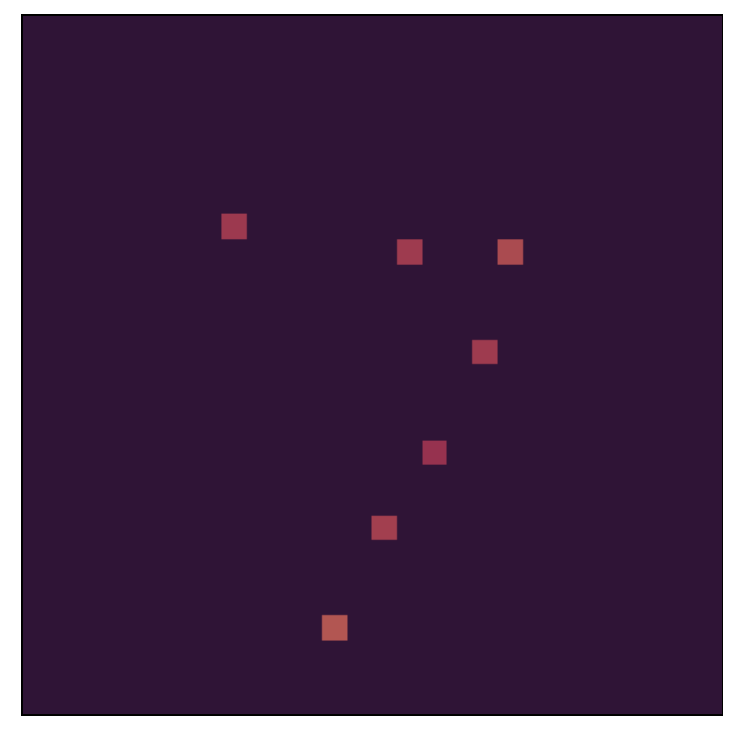}\label{subfig:extrema}}
	\caption{Visualization of the activation maps of five activation functions (Identity, ReLU, top-k absolutes, Extrema-Pool indices and Extrema) for 1D and 2D input in the top and bottom row respectively.
	The 1D input to the activation functions is denoted with the continuous transparent green line using an example from the UCI dataset.
	The output of each activation function is denoted with the cyan stem lines with blue markers.
	The 2D example depicts only the output of the activation functions using an example from the MNIST dataset.
	}\label{fig:activationfunctions}
\end{figure*}

Initially, bounded functions such as $sigmoid$ and $\tanh$ were used, however besides producing dense activation maps they also present the vanishing gradients problem~\cite{bengio1994learning}.
Rectified Linear Units (ReLUs) were later proposed~\cite{glorot2011deep, nair2010rectified} as an activation function that solves the vanishing gradients problem and increases the sparsity of the activation maps.
Although ReLU creates exact zeros (unlike its predecessors $sigmoid$ and $\tanh$), its activation map consists of sparsely separated but still dense areas (Fig.~\ref{fig:activationfunctions}\subref{subfig:relu}) instead of sparse spikes.
The same applies for generalizations of ReLU, such as Parametric ReLU~\cite{he2015delving} and Maxout~\cite{goodfellow2013maxout}.
Recently, in $k$-Sparse Autoencoders~\cite{makhzani2013k} the authors used an activation function that applies thresholding until the $k$ most active activations remain, however this non-linearity covers a limited area of the activation map by creating sparsely disconnected dense areas (Fig.~\ref{fig:activationfunctions}\subref{subfig:topkabsolutes}), similar to the ReLU case.

Moreover activation functions that produce continuous valued activation maps (such as ReLU) are less biologically plausible, because biological neurons rarely are in their maximum saturation regime~\cite{bush1996inhibition} and use spikes to communicate instead of continuous values~\cite{bengio2015towards}.
Previous literature has also demonstrated the increased biological plausibility of sparseness in artificial neural networks~\cite{rehn2007network}.
Spike-like sparsity on activation maps has been thoroughly researched on the more biologically plausible rate-based network models~\cite{heiberg2018firing}, but it has not been thoroughly explored as a design option for activation functions combined with convolutional filters.

The increased number of weights and non-zero activations make DNNs more complex, and thus more difficult to use in problems that require corresponding causality of the output with a specific set of neurons.
The majority of domains where machine learning is applied, including critical areas such as healthcare~\cite{bizopoulos2019deep}, require models to be interpretable and explainable before considering them as a solution.
Although these properties can be increased using sensitivity analysis~\cite{simonyan2013deep}, deconvolution methods~\cite{zeiler2014visualizing}, Layerwise-Relevance Propagation~\cite{bach2015pixel} and Local-Interpretable Model agnostic Explanations~\cite{ribeiro2016should} it would be preferable to have self-interpretable models.

Moreover, considering that DNNs learn to represent data using the combined set of trainable weights and non-zero activations during the feed-forward pass, an interesting question arises:
\\\\
\indent\textit{What are the implications of trading off the reconstruction error of the representations with their compression ratio w.r.t to the original data?}
\\

\begin{figure*}[!t]
	\subfloat[1D SAN]{\begin{tikzpicture}[]
		\begin{scope}[tdplot_main_coords, canvas is yz plane at x=-0.5,xscale=-1, transform shape]
			\node[opacity=0] at (0, 0)(input){\includegraphics[scale=\figscale]{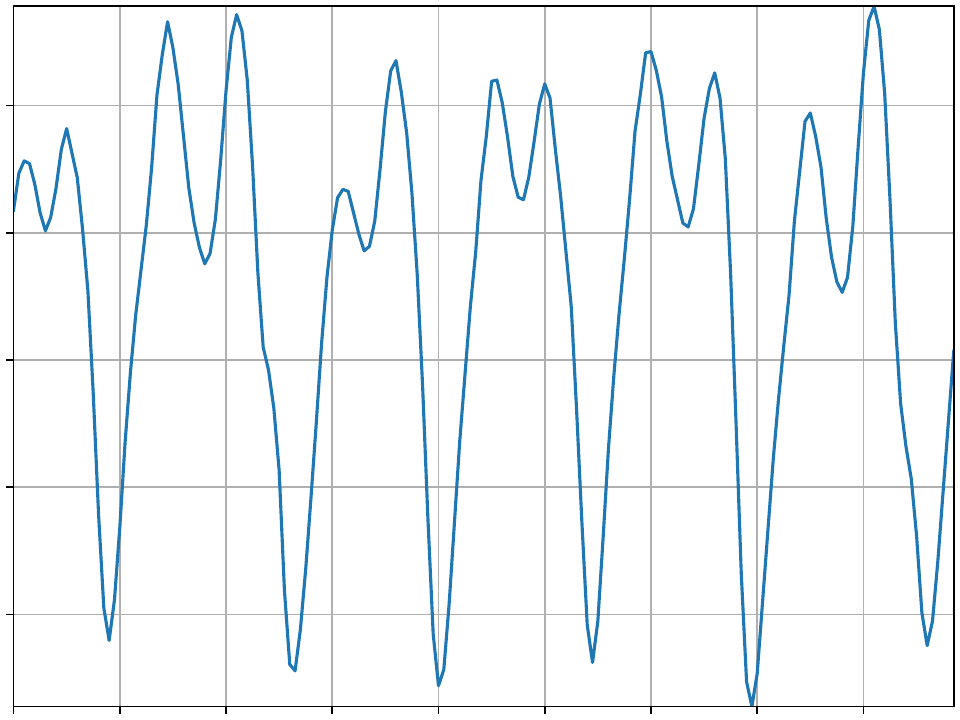}};
			\node[opacity=0, draw, right=0.5cm of input, circle] (loss){$\mathcal{L}$};
			\node[opacity=0, right=0.5cm of loss] (reconstructed){\includegraphics[scale=\figscale]{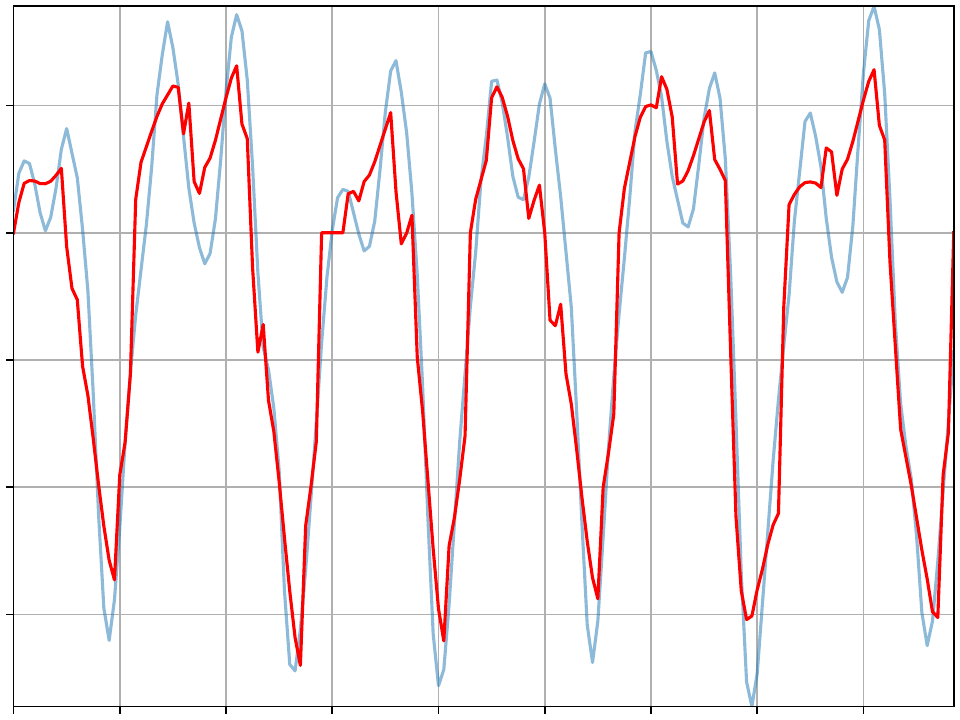}};
			\node[opacity=0, draw, below=0.5cm of reconstructed, circle] (plus){$+$};
			\node[opacity=0.8, below=0.5cm of plus] (reconstruction){\includegraphics[scale=\figscale]{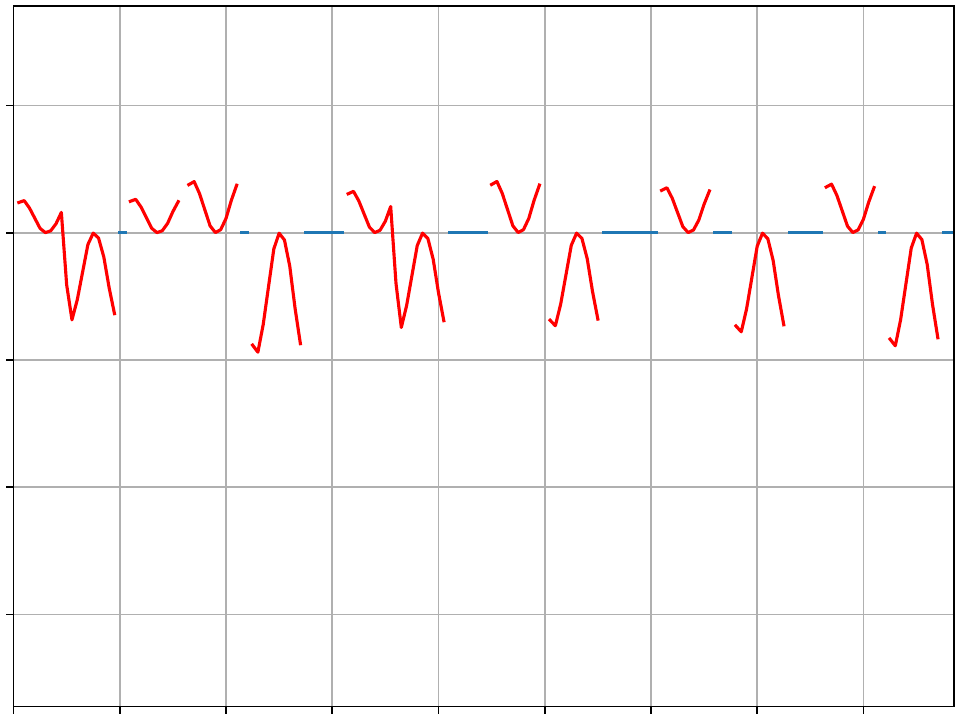}};
			\node[opacity=0, draw, below=0.5cm of reconstruction, circle] (conv2){$\ast$};
			\node[] at (4, -3){$\bm{r}^{(0)}$};
			\node[opacity=0.8, below=0.5cm of conv2] (extrema){\includegraphics[scale=\figscale]{python/tmp/UCI-epilepsy-extrema-1d-2-activations-0.pdf}};
			\node[opacity=0, draw, left=0.55cm of extrema, circle] (phi){$\phi$};
			\node[] at (4, -6.3){$\bm{\alpha}^{(0)}$};
			\node[opacity=0.8, left=0.55cm of phi] (similarity){\includegraphics[scale=\figscale]{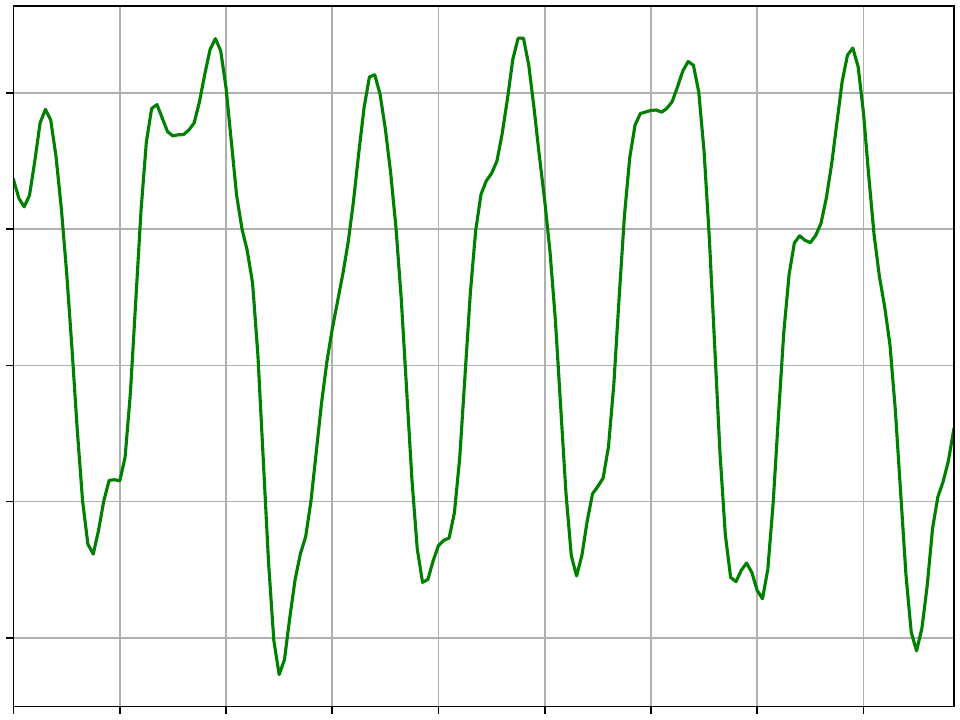}};
			\node[] at (0.1, -6.3){$\bm{s}^{(0)}$};
			\node[left=1.5cm of conv2] (kernel){\includegraphics[scale=0.1]{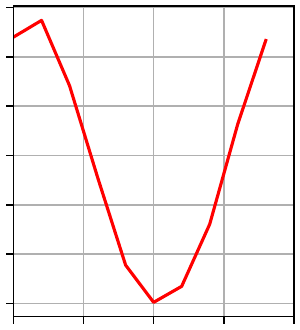}};
			\node[] at (2, -4.7){\tiny $\bm{w}^{(0)}$};
			\node[opacity=0, draw, above=0.5cm of similarity, circle] (conv1){$\ast$};
		\end{scope}
		\begin{scope}[tdplot_main_coords, canvas is yz plane at x=0,xscale=-1, transform shape]
			\node[] at (0, 0)(input){\includegraphics[scale=\figscale]{python/tmp/UCI-epilepsy-extrema-1d-2-signal.pdf}};
			\node[] at (0.1, 0.5){$\bm{x}$};
			\node[draw, right=0.5cm of input, circle] (loss){$\mathcal{L}$};
			\node[right=0.5cm of loss] (reconstructed){\includegraphics[scale=\figscale]{python/tmp/UCI-epilepsy-extrema-1d-2-reconstructed.pdf}};
			\node[] at (4, 0.5){$\hat{\bm{x}}$};
			\node[draw, below=0.5cm of reconstructed, circle] (plus){$+$};
			\node[opacity=0, below=0.5cm of plus] (reconstruction){\includegraphics[scale=\figscale]{python/tmp/UCI-epilepsy-extrema-1d-2-reconstruction-0.pdf}};
			\node[draw, below=0.5cm of reconstruction, circle] (conv2){$\ast$};
			\node[opacity=0, below=0.5cm of conv2] (extrema){\includegraphics[scale=\figscale]{python/tmp/UCI-epilepsy-extrema-1d-2-activations-0.pdf}};
			\node[draw, left=0.55cm of extrema, circle, inner sep=2pt] (phi){$\phi$};
			\node[opacity=0, left=0.55cm of phi] (similarity){\includegraphics[scale=\figscale]{python/tmp/UCI-epilepsy-extrema-1d-2-similarity-0.pdf}};
			\node[opacity=0, left=1.5cm of conv2] (kernel){\includegraphics[scale=0.1]{python/tmp/UCI-epilepsy-extrema-1d-2-kernel-0.pdf}};
			\node[draw, above=0.5cm of similarity, circle] (conv1){$\ast$};
			\draw[->](input) -- node{} (conv1);
			\draw[->](conv1) -- node{} (similarity);
			\draw[->](similarity) -- node{} (phi);
			\draw[->](phi) -- node{} (extrema);
			\draw[->](extrema) -- node{} (conv2);
			\draw[->](conv2) -- node{} (reconstruction);
			\draw[->](reconstruction) -- node{} (plus);
			\draw[->](plus) -- node{} (reconstructed);
			\draw[->](kernel) -- node{} (conv1);
			\draw[->](kernel) -- node{} (conv2);
			\draw[->](input) -- node{} (loss);
			\draw[->](reconstructed) -- node{} (loss);
		\end{scope}
		\begin{scope}[tdplot_main_coords, canvas is yz plane at x=0.5,xscale=-1, transform shape]
			\node[opacity=0] at (0, 0)(input){\includegraphics[scale=\figscale]{python/tmp/UCI-epilepsy-extrema-1d-2-signal.pdf}};
			\node[opacity=0, draw, right=0.5cm of input, circle] (loss){$\mathcal{L}$};
			\node[opacity=0, right=0.5cm of loss] (reconstructed){\includegraphics[scale=\figscale]{python/tmp/UCI-epilepsy-extrema-1d-2-reconstructed.pdf}};
			\node[opacity=0, draw, below=0.5cm of reconstructed, circle] (plus){$+$};
			\node[opacity=0.8, below=0.5cm of plus] (reconstruction){\includegraphics[scale=\figscale]{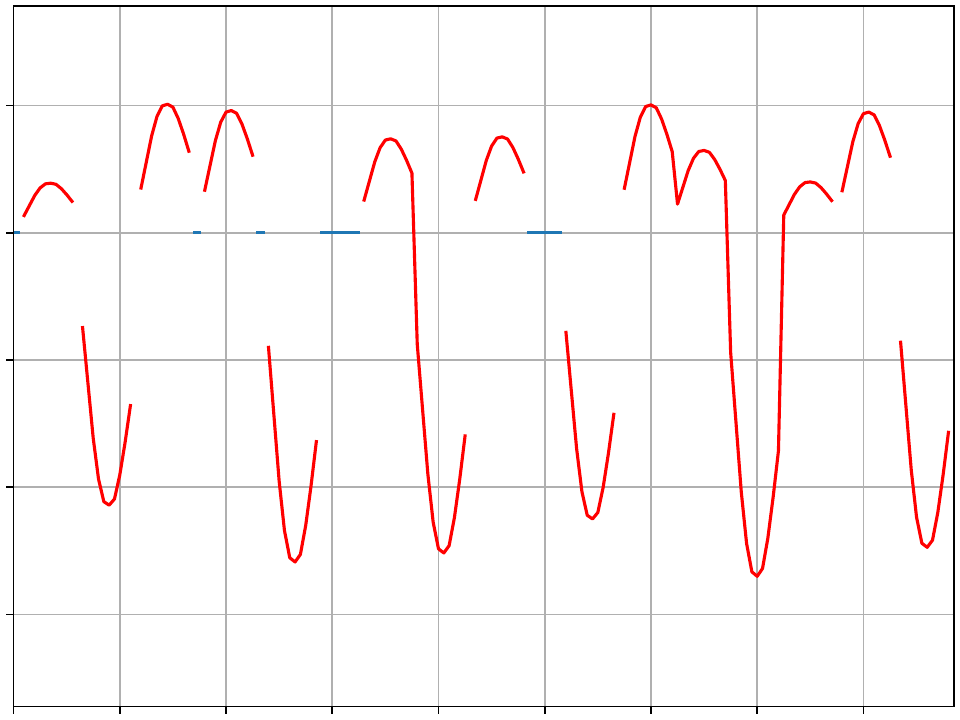}};
			\node[] at (4, -3){$\bm{r}^{(1)}$};
			\node[opacity=0, draw, below=0.5cm of reconstruction, circle] (conv2){$\ast$};
			\node[opacity=0.8, below=0.5cm of conv2] (extrema){\includegraphics[scale=\figscale]{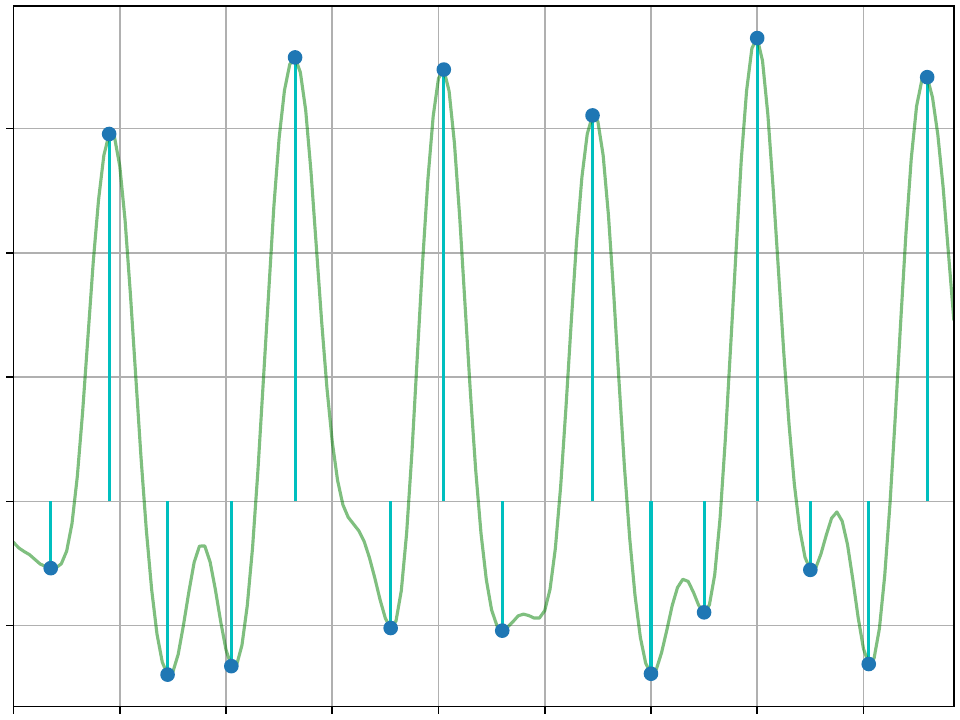}};
			\node[] at (4, -6.3){$\bm{\alpha}^{(1)}$};
			\node[opacity=0, draw, left=0.55cm of extrema, circle] (phi){$\phi$};
			\node[opacity=0.8, left=0.55cm of phi] (similarity){\includegraphics[scale=\figscale]{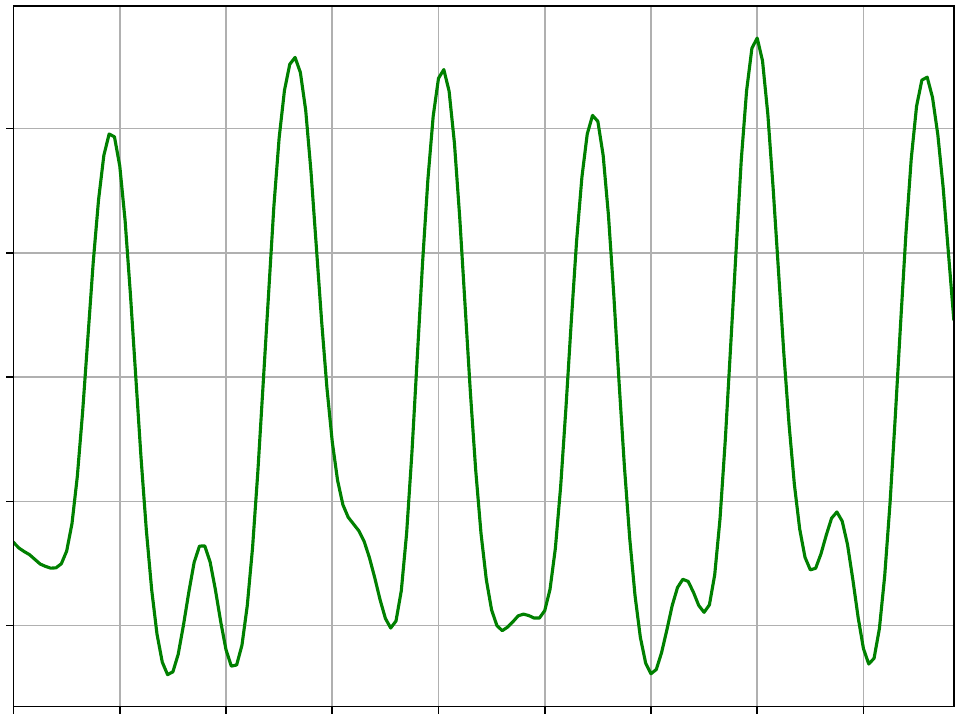}};
			\node[] at (0.1, -6.3){$\bm{s}^{(1)}$};
			\node[left=1.5cm of conv2] (kernel){\includegraphics[scale=0.1]{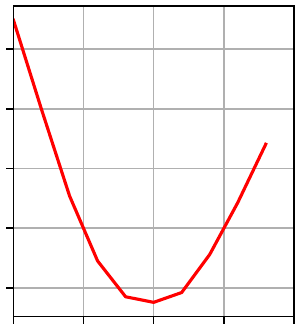}};
			\node[] at (2, -4.7){\tiny $\bm{w}^{(1)}$};
			\node[opacity=0, draw, above=0.5cm of similarity, circle] (conv1){$\ast$};
		\end{scope}
	\end{tikzpicture}
	}
	\qquad
	\qquad
	\qquad
	\qquad
	\qquad
	\qquad
	\subfloat[2D SAN]{\begin{tikzpicture}[]
		\begin{scope}[tdplot_main_coords, canvas is yz plane at x=-0.5,xscale=-1, transform shape]
			\node[opacity=0] at (0, 0)(input){\includegraphics[scale=\figscale]{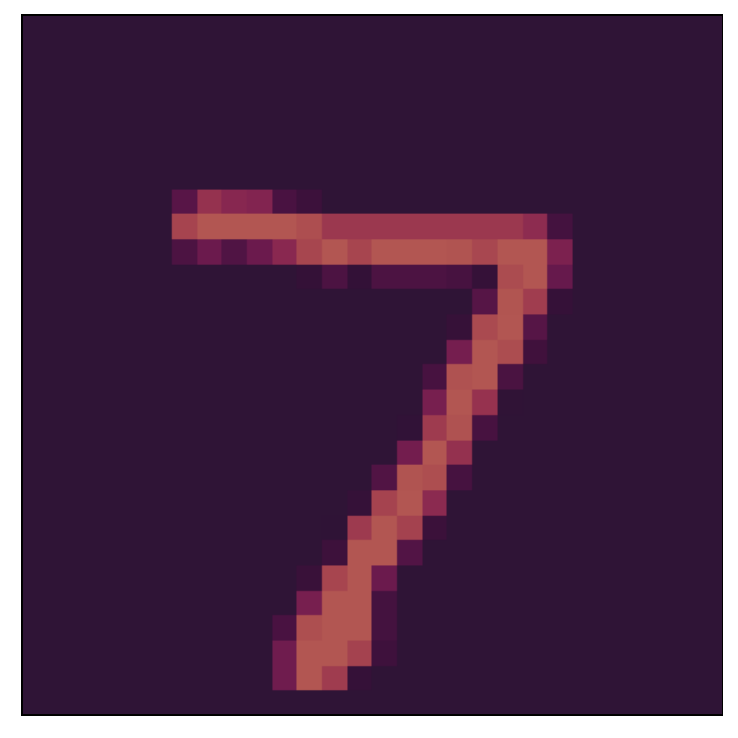}};
			\node[opacity=0, draw, right=0.5cm of input, circle] (loss){$\mathcal{L}$};
			\node[opacity=0, right=0.5cm of loss] (reconstructed){\includegraphics[scale=\figscale]{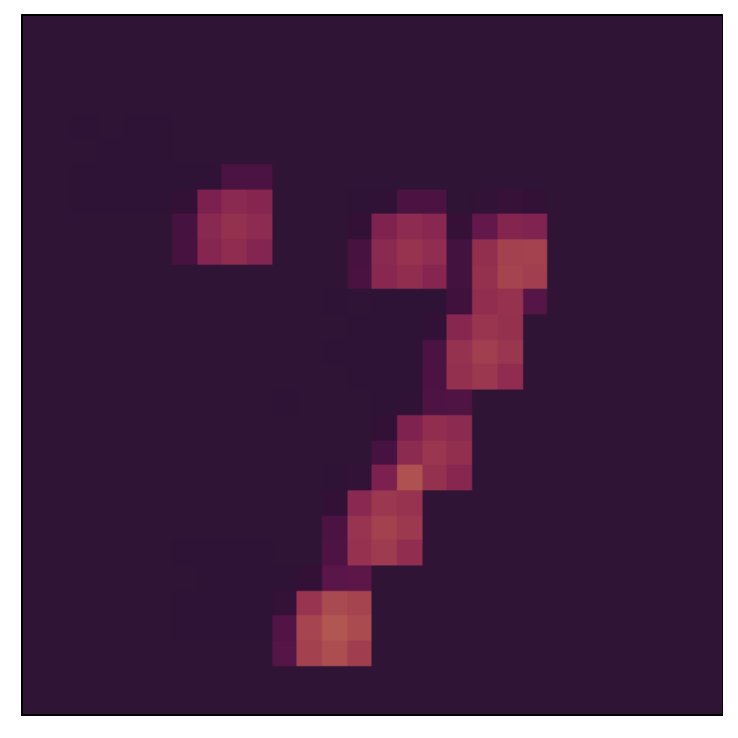}};
			\node[opacity=0, draw, below=0.5cm of reconstructed, circle] (plus){$+$};
			\node[opacity=0.8, below=0.5cm of plus] (reconstruction){\includegraphics[scale=\figscale]{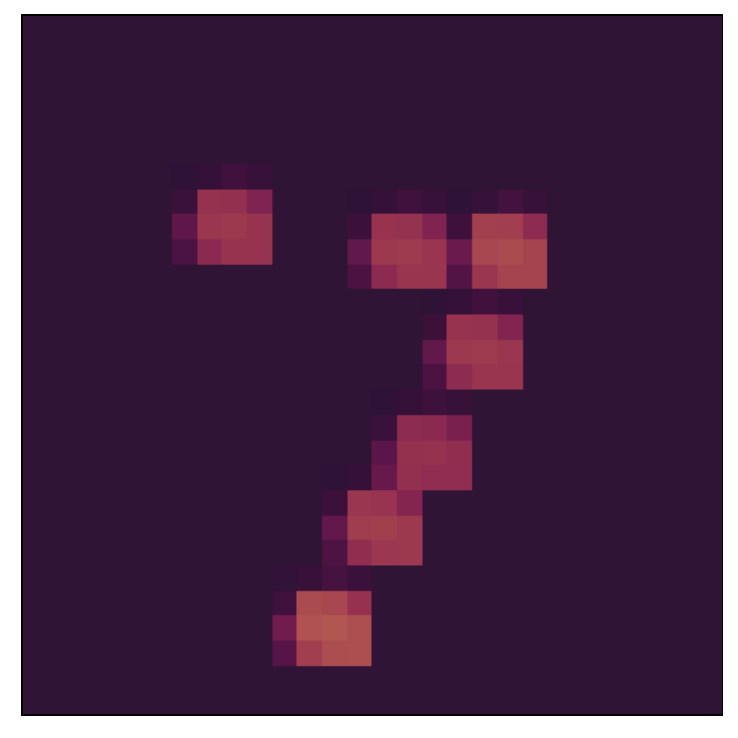}};
			\node[opacity=0, draw, below=0.5cm of reconstruction, circle] (conv2){$\ast$};
			\node[opacity=0.8, below=0.5cm of conv2] (extrema){\includegraphics[scale=\figscale]{python/tmp/MNIST-extrema-2d-2-activations-0.pdf}};
			\node[opacity=0, draw, left=0.55cm of extrema, circle] (phi){$\phi$};
			\node[opacity=0.8, left=0.55cm of phi] (similarity){\includegraphics[scale=\figscale]{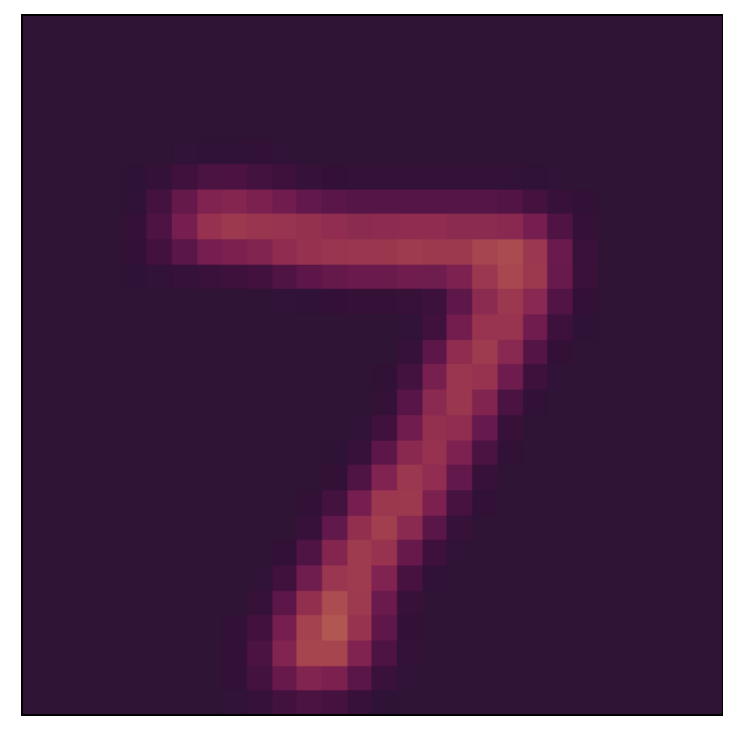}};
			\node[opacity=0.8, left=1.2cm of conv2] (kernel){\includegraphics[scale=0.1]{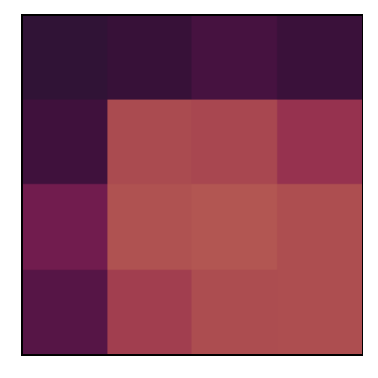}};
			\node[opacity=0, draw, above=0.5cm of similarity, circle] (conv1){$\ast$};
		\end{scope}
		\begin{scope}[tdplot_main_coords, canvas is yz plane at x=0,xscale=-1, transform shape]
			\node[] at (0, 0)(input){\includegraphics[scale=\figscale]{python/tmp/MNIST-extrema-2d-2-signal.pdf}};
			\node[draw, right=0.5cm of input, circle] (loss){$\mathcal{L}$};
			\node[right=0.5cm of loss] (reconstructed){\includegraphics[scale=\figscale]{python/tmp/MNIST-extrema-2d-2-reconstructed.pdf}};
			\node[draw, below=0.5cm of reconstructed, circle] (plus){$+$};
			\node[opacity=0, below=0.5cm of plus] (reconstruction){\includegraphics[scale=\figscale]{python/tmp/MNIST-extrema-2d-2-reconstruction-0.pdf}};
			\node[draw, below=0.5cm of reconstruction, circle] (conv2){$\ast$};
			\node[opacity=0, below=0.5cm of conv2] (extrema){\includegraphics[scale=\figscale]{python/tmp/MNIST-extrema-2d-2-activations-0.pdf}};
			\node[draw, left=0.55cm of extrema, circle, inner sep=2pt] (phi){$\phi$};
			\node[opacity=0, left=0.55cm of phi] (similarity){\includegraphics[scale=\figscale]{python/tmp/MNIST-extrema-2d-2-similarity-0.pdf}};
			\node[opacity=0, left=1.2cm of conv2] (kernel){\includegraphics[scale=0.1]{python/tmp/MNIST-extrema-2d-2-kernel-0.pdf}};
			\node[draw, above=0.5cm of similarity, circle] (conv1){$\ast$};
			\draw[->](input) -- node{} (conv1);
			\draw[->](conv1) -- node{} (similarity);
			\draw[->](similarity) -- node{} (phi);
			\draw[->](phi) -- node{} (extrema);
			\draw[->](extrema) -- node{} (conv2);
			\draw[->](conv2) -- node{} (reconstruction);
			\draw[->](reconstruction) -- node{} (plus);
			\draw[->](plus) -- node{} (reconstructed);
			\draw[->](kernel) -- node{} (conv1);
			\draw[->](kernel) -- node{} (conv2);
			\draw[->](input) -- node{} (loss);
			\draw[->](reconstructed) -- node{} (loss);
		\end{scope}
		\begin{scope}[tdplot_main_coords, canvas is yz plane at x=0.5,xscale=-1, transform shape]
			\node[opacity=0] at (0, 0)(input){\includegraphics[scale=\figscale]{python/tmp/MNIST-extrema-2d-2-signal.pdf}};
			\node[opacity=0, draw, right=0.5cm of input, circle] (loss){$\mathcal{L}$};
			\node[opacity=0, right=0.5cm of loss] (reconstructed){\includegraphics[scale=\figscale]{python/tmp/MNIST-extrema-2d-2-reconstructed.pdf}};
			\node[opacity=0, draw, below=0.5cm of reconstructed, circle] (plus){$+$};
			\node[opacity=0.8, below=0.5cm of plus] (reconstruction){\includegraphics[scale=\figscale]{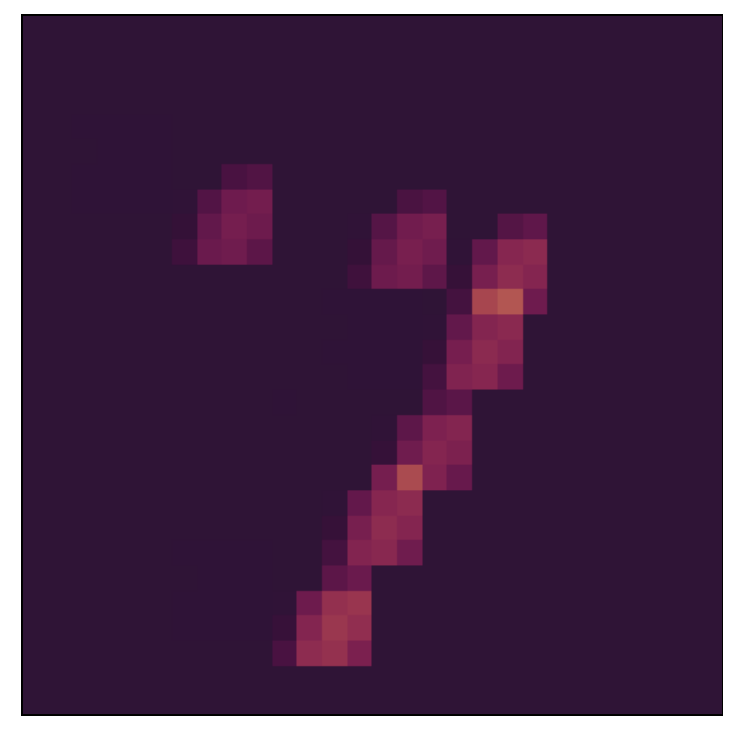}};
			\node[opacity=0, draw, below=0.5cm of reconstruction, circle] (conv2){$\ast$};
			\node[opacity=0.8, below=0.5cm of conv2] (extrema){\includegraphics[scale=\figscale]{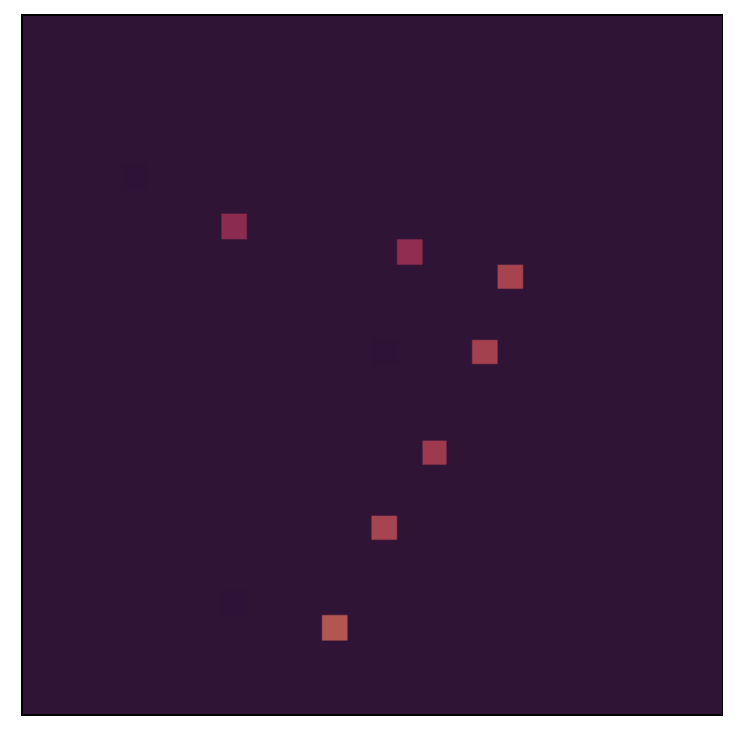}};
			\node[opacity=0, draw, left=0.55cm of extrema, circle] (phi){$\phi$};
			\node[opacity=0.8, left=0.55cm of phi] (similarity){\includegraphics[scale=\figscale]{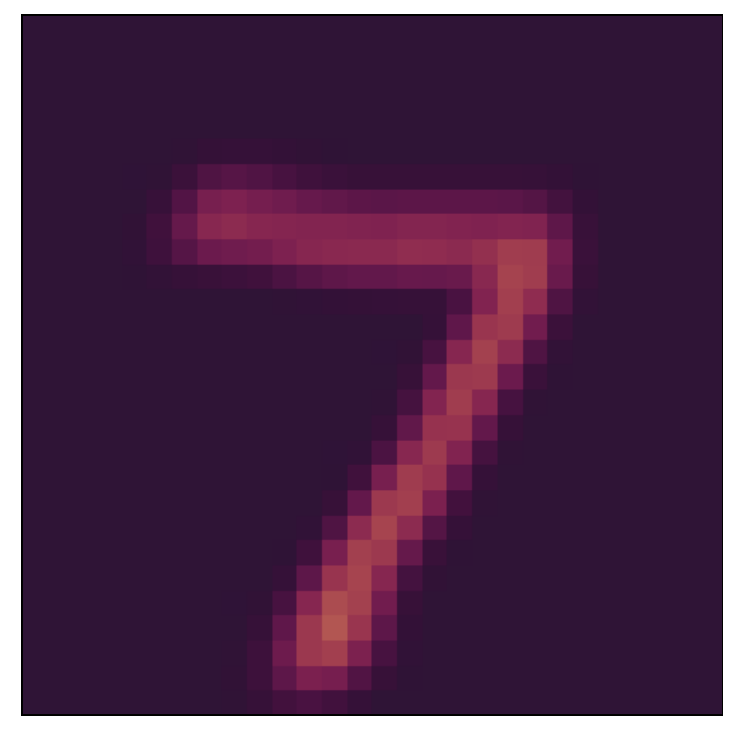}};
			\node[opacity=0.8, left=1.2cm of conv2] (kernel){\includegraphics[scale=0.1]{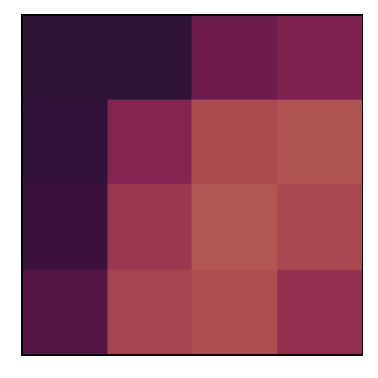}};
			\node[opacity=0, draw, above=0.5cm of similarity, circle] (conv1){$\ast$};
		\end{scope}
	\end{tikzpicture}
	}
	\caption{Diagrams of the feed-forward pass of an 1D and 2D SAN with two kernels for random examples from the test dataset of UCI epilepsy database and MNIST respectively.
	The figures depict intermediate representations; $\bm{x}$ denotes the input signal (blue line), $\bm{w}^{(i)}$ denotes the kernels (red line), $\bm{s}^{(i)}$ denotes the similarity matrices (green line), $\bm{\alpha}^{(i)}$ denotes the activation maps (cyan stem lines with blue markers), $\bm{r}^{(i)}$ denotes the partial reconstruction from each $\bm{w}^{(i)}$ and $\hat{\bm{x}}$ denotes the reconstructed input (red line).
	Placed for comparison, the transparent green line in $\bm{\alpha}^{(i)}$ denotes the corresponding $\bm{s}^{(i)}$ and the transparent blue line in $\hat{\bm{x}}$ denotes the input $\bm{x}$.
	The exponent $i=0,1$ corresponds to the first and second kernel and the intermediate representations respectively.
	The circles denote operations; $\mathcal{L}$ denotes the loss function, $\phi$ denotes the sparse activation function, $\ast$ the convolution operation and $+$ the plus operation.
	All operations are performed separate for each $\bm{w}^{(i)}$ however for visual clarity we only depict one operation for each step.
	Shades of red and blue in the 2D example represent positive and negative values respectively.
	The Extrema activation function was used for both examples.
	}\label{fig:sans}
\end{figure*}

Previous work by Blier et al.~\cite{blier2018description} demonstrated the ability of DNNs to losslessly compress the input data and the weights, but without considering the number of non-zero activations.
In this work we relax the lossless requirement and also consider neural networks purely as function approximators instead of probabilist models.
The contributions of this paper are the following proposals:
\begin{itemize}
	\item The $\varphi$ metric that evaluates unsupervised models based on how compressed their learned representations are w.r.t the original data and how accurate their reconstruction is.
	\item Sparsely Activated Networks (SANs) (Fig.~\ref{fig:sans}) in which spike-like sparsity is enforced in the activation map (Fig.~\ref{fig:activationfunctions}\subref{subfig:extremapoolindices} and \subref{subfig:extrema}) through the use of a sparse activation function.
\end{itemize}

In Section~\ref{sec:flithos} we define the $\varphi$ metric, then in Section~\ref{sec:sans} we define the five tested activation functions along with the architecture and training procedure of SANs, in Section~\ref{sec:experiments} we experiment SANs on the Physionet~\cite{goldberger2000physiobank}, UCI-epilepsy~\cite{andrzejak2001indications}, MNIST~\cite{lecun1998gradient} and FMNIST~\cite{xiao2017fashion} databases and provide visualizations of the intermediate representations and results.
In Section~\ref{sec:discussion} we discuss the findings of the experiments and the limitations of SANs and finally in Section~\ref{sec:conclusions} we present the concluding remarks and the future work.

\section{$\varphi$ Metric}\label{sec:flithos}
Let $M$ be a model with $q$ kernels each one with $m^{(i)}$ samples and a reconstruction loss function $\mathcal{L}$ for which we have:
\begin{equation}
	\label{eq:model}
	M: \bm{x} \longmapsto \hat{\bm{x}}
\end{equation}

\noindent
, where $\bm{x} \in \mathbb{R}^n$ is the input vector and $\hat{\bm{x}}$ is a reconstruction of $\bm{x}$.
For the definition of $\varphi$ metric we use a neural network model that consists of convolutional filters, however this can be easily generalized to other architectures.

The $\varphi$ metric evaluates a model based on two concepts: its verbosity and its accuracy.
Verbosity in neural networks can be perceived as inversely proportional to the compression ratio of the representations.
We calculate the number of weights $W$ of $M$ as follows:
\begin{equation}
	\label{eq:numberofweights}
	W = \sum\limits_{i=1}^q m^{(i)}
\end{equation}

We also calculate the number of non-zero activations $A$ of $M$ for input $\bm{x} \in \mathbb{R}^n$ as:
\begin{equation}
	\label{eq:numberofactivations}
	A_{\bm{x}} = \sum\limits_{i=1}^q \Big\lVert\bm{\alpha}^{(i)}\Big\rVert_0
\end{equation}

\noindent
, where $\lVert \cdot \rVert_0$ denotes the $\ell_0$ pseudo-norm and $\bm{\alpha}^{(i)}$ is the activation map of the $i^{th}$ kernel.
Then using Equations~\ref{eq:numberofweights} and~\ref{eq:numberofactivations} we define the compression ratio $CR$ of $\bm{x}$ w.r.t $M$ as:
\begin{equation}
	\label{eq:compressionratio}
	CR = \frac{n}{W + (\dim(\bm{x}) + 1)A_{\bm{x}}}
\end{equation}

\noindent
, where $\dim$ denotes dimensionality and $n$ was previously defined as the cardinality of $\bm{x}$.
The reason that we multiply the dimensionality of $\bm{x}$ with the number of activations $A_{\bm{x}}$ is that we need to consider the spatial position of each non-zero activation in addition to its amplitude to reconstruct $\bm{x}$.
Moreover, using that definition of $CR$, there is a desirable trade-off between using a larger kernel with less instances and a smaller kernel with more instances based on which kernel size minimizes the $CR$.
This definition of $CR$ allows us to set a base of reference for $CR=1$ for models that their representational capacity is equal with the number of samples of the input.

Regarding the accuracy, we define the normalized reconstruction loss as follows:
\begin{equation}
	\label{eq:normalizedreconstrucionloss}
	\tilde{\mathcal{L}}(\hat{\bm{x}},\bm{x}) = \frac{\mathcal{L}(\hat{\bm{x}},\bm{x})}{\mathcal{L}(0,\bm{x})}
\end{equation}

This definition of $\tilde{\mathcal{L}}$ allows us to set a base of reference for $\tilde{\mathcal{L}}$ in cases when the reconstruction the model performs is equivalent with a model that performs constant reconstructions independent of the input.
Finally using Equations~\ref{eq:compressionratio} and~\ref{eq:normalizedreconstrucionloss} we define the $\varphi$\footnote{The use of the symbol $\varphi$ comes from the first character of the combined greek word {\textgreekfont{φλύθος}} = {\textgreekfont{φλύ+θος}} (flithos: pronounced as flee-thos). It consists of the first part of the word {\textgreekfont{φλύ-αρος}} (meaning verbose) and the second part of {\textgreekfont{λά-θος}} (meaning wrong). {\textgreekfont{Φλύθος}} is literally defined as: \textit{Giving inaccurate information using many words; the state of being wrong and wordy at the same time}.} metric of $\bm{x}$ w.r.t $M$ as follows:
\begin{equation}
	\label{eq:flithos}
	\varphi=\lVert(CR^{-1},\tilde{\mathcal{L}}(\hat{\bm{x}},\bm{x}))\rVert_2
\end{equation}

\noindent
, where $\lVert \cdot \rVert_2$ denotes the $\ell_2$ euclidean-norm.
The rationale behind defining $\varphi$ is to satisfy the need for a unified metric that takes in consideration both the `verbosity' of a model along with its `accuracy'.

Regarding hyperparameter selection we also define the mean $\varphi$ of a dataset or a mini-batch w.r.t to $M$ as:
\begin{equation}
	\label{eq:meanflithos}
	\bar\varphi = \frac{1}{l}\sum\limits_{j=1}^l \varphi^{(j)}
\end{equation}

\noindent
, where $l$ is the number of observations in the dataset or the batch size.

$\bar\varphi$ is non-differentiable due to the presence of the $\ell_0$ pseudo-norm in Eq.~\ref{eq:numberofactivations}.
A way to overcome this is using $\mathcal{L}$ as the differentiable optimization function during training and $\bar\varphi$ as the metric for model selection during validation on which hyperparameter value decisions (such as kernel size) are made.

\section{Sparsely Activated Networks}\label{sec:sans}

\subsection{Sparse Activation Functions}\label{sec:safs}
In this subsection we define five activation functions $\phi$ and their corresponding sparsity density parameter $d^{(i)}$ for which we have:
\begin{equation}
	\label{eq:phi}
	\phi: s \longmapsto \alpha
\end{equation}

We choose values for $d^{(i)}$ for each activation function in such as way, to approximately have the same number of activations for fair comparison of the sparse activation functions.

\subsubsection{Identity}\label{sec:identity}
$\phi = \mathds{1}$.
The Identity activation function serves as a baseline and does not change its input as shown in Fig.~\ref{fig:activationfunctions}\subref{subfig:identity}.
For this case $d^{(i)}$ does not apply.

\subsubsection{ReLU}\label{sec:relu}
$\phi = ReLU(s)$.
The ReLU activation function produces sparsely disconnected but internally dense areas as shown in Fig.~\ref{fig:activationfunctions}\subref{subfig:relu} instead of sparse spikes.
For this case $d^{(i)}$ does not apply.

\subsubsection{top-k absolutes}\label{sec:topkabsolutes}
The top-k absolutes function (defined at Algorithm~\ref{alg:topkabsolutes}) keeps the indices of $k$ activations with the largest absolute value and zeros out the rest, where $1 \le k < n \in \mathbb{N}$.
We set $d^{(i)} = k$, where $k = \lfloor n/m \rfloor^{\dim(\bm{x})}$.
Top-k absolutes is sparser than ReLU but some extrema are overactivated related to some others that are not activated at all, as shown in Fig.~\ref{fig:activationfunctions}\subref{subfig:topkabsolutes}.

\begin{algorithm}[H]
	\caption{top-k absolutes}\label{alg:topkabsolutes}
	\begin{algorithmic}[1]
		\renewcommand{\algorithmicrequire}{\textbf{Input:}}
		\renewcommand{\algorithmicensure}{\textbf{Output:}}
		\REQUIRE{$s$, $k$}
		\ENSURE{$\alpha$}
		\STATE{$\alpha(i) \leftarrow 0, i=1\ldots card(s)$}
		\STATE{$e \leftarrow topk(\lvert s\rvert, k)$}
		\\\textit{\scriptsize \# e denotes the vector with the extrema indices}
		\FOR{$i$ = 0 to $card(s)$}
		\STATE{$\alpha(e(i)) \leftarrow s(e(i))$}
		\ENDFOR{}
		\RETURN{$\alpha$}
	\end{algorithmic}
\end{algorithm}

\subsubsection{Extrema-Pool indices}\label{sec:extremapoolindices}
The Extrema-Pool indices activation function (defined at Algorithm~\ref{alg:extremapoolindices}) keeps only the index of the activation with the maximum absolute amplitude from each region outlined by a grid as granular as the kernel size $m^{(i)}$ and zeros out the rest.
It consists of a max-pooling layer followed by a max-unpooling layer with the same parameters while the sparsity parameter $d^{(i)}$ in this case is set $d^{(i)} = m^{(i)} < n \in \mathbb{N}$.
This activation function creates sparser activation maps than top-k absolutes however in cases where the pool grid is near a peak or valley this region is activated twice (as shown in Fig.~\ref{fig:activationfunctions}\subref{subfig:extremapoolindices}).

\begin{algorithm}[H]
	\caption{Extrema-Pool indices}\label{alg:extremapoolindices}
	\begin{algorithmic}[1]
		\renewcommand{\algorithmicrequire}{\textbf{Input:}}
		\renewcommand{\algorithmicensure}{\textbf{Output:}}
		\REQUIRE{$s$, $m$}
		\ENSURE{$\alpha$}
		\STATE{$e \leftarrow maxpool(\lvert s\rvert, m)$}
		\STATE{$\alpha \leftarrow maxunpool(s(e), e, m)$}
		\RETURN{$\alpha$}
	\end{algorithmic}
\end{algorithm}

\subsubsection{Extrema}\label{sec:extrema}
The Extrema activation function (defined at Algorithm~\ref{alg:extrema}) detects candidate extrema using zero crossing of the first derivative, then sorts them in an descending order and gradually eliminates those extrema that have less absolute amplitude than a neighboring extrema within a predefined minimum extrema distance ($med$).
Imposing a $med$ on the extrema detection algorithm makes $\bm{\alpha}$ sparser than the previous cases and solves the problem of double extrema activations that Extrema-Pool indices have (as shown in Fig.~\ref{fig:activationfunctions}\subref{subfig:extrema}).
The sparsity parameter in this case is set $d^{(i)} = med$, where $1 \le med < n \in \mathbb{N}$ is the minimum extrema distance.
We set $med = m^{(i)}$ for utilizing fair comparison between the sparse activation functions.
Specifically for Extrema activation function we introduce a `border tolerance' parameter to allow neuron activation within another neuron activated area.

\begin{algorithm}[H]
	\caption{Extrema detection with minimum extrema distance $med$}\label{alg:extrema}
	\begin{algorithmic}[1]
		\renewcommand{\algorithmicrequire}{\textbf{Input:}}
		\renewcommand{\algorithmicensure}{\textbf{Output:}}
		\REQUIRE{$s$, $med$}
		\ENSURE{$\alpha$}
		\STATE{$peaks \leftarrow \left(\frac{d s}{d t}^+ \geq 0\right) \land \left(\frac{d s}{d t}^- < 0\right)$}
		\STATE{$valleys \leftarrow \left(\frac{d s}{d t}^+ < 0\right) \land \left(\frac{d s}{d t}^- \geq 0\right)$}
		\\\textit{\scriptsize \# + and - denote one sample padding to the right and left respectively}
		\STATE{$z \leftarrow peaks \lor valleys$}
		\STATE{$e \leftarrow z > 0$}
		\STATE{$e_{ind} \leftarrow sort(z)$}
		\STATE{$e_{sorted} \leftarrow e(e_{ind})$}
		\STATE{$e_{seco}(i) \leftarrow 0, i=1\ldots card(s)$}
		\\\textit{\scriptsize \# $e_{seco}$ denotes a boolean vector for the secondary extrema indices}
		\FOR{$i$ = 0 to $card(s)$}
		\IF{$\lnot e_{seco}(i)$}
		\STATE{$e_r \leftarrow e \geq e_{ind}(i) - med$}
		\STATE{$e_l \leftarrow e \leq e_{ind}(i) + med$}
		\STATE{$e_m \leftarrow e_r \land e_l$}
		\STATE{$e_{seco} \leftarrow e_{seco} \lor e_m$}
		\STATE{$e_{seco}(i) \leftarrow 0$}
		\ENDIF{}
		\ENDFOR{}
		\STATE{$\alpha_{ind} \leftarrow e_{sorted}(\lnot e_{seco})$}
		\\\textit{\scriptsize \# $\alpha_{ind}$ denotes the indices of $\alpha$ with non-zero value}
		\STATE{$\alpha(i) \leftarrow 0, i=1\ldots card(s)$}
		\FOR{$i$ = 0 to $card(s)$}
		\STATE{$\alpha(\alpha_{ind}(i)) \leftarrow s(\alpha_{ind}(i))$}
		\ENDFOR{}
		\RETURN{$\alpha$}
	\end{algorithmic}
\end{algorithm}

\subsection{SAN Architecture/Training}

Let $\bm{x} \in \mathbb{R}^n$ be a single input data, however the following can be trivially generalized for batch inputs with different cardinalities.
Let $\bm{w}^{(i)} \in \mathbb{R}^{m^{(i)}}$ the weight matrix of the $i^{th}$ kernel, that are initialized using a normal distribution with mean $\mu$ and standard deviation $\sigma$:
\begin{equation}
	\label{eq:weightinitialization}
	\bm{w}^{(i)} \sim \mathcal{N}(\mu, \sigma)
\end{equation}

\noindent
, where $0\le i < q \in \mathbb{N}$ is the number of kernels.

First we calculate the similarity matrices\footnote{Previous literature refers it as the `hidden variable' but we use a more direct naming that suits the context of this paper.} $\bm{s}^{(i)}$ for each of the weight matrices $\bm{w}^{(i)}$:
\begin{equation}
	\label{eq:similarity}
	\bm{s}^{(i)} = \bm{x} * \bm{w}^{(i)}
\end{equation}

\noindent
, where $*$ is the convolution\footnote{We use convolution instead of cross-correlation only as a matter of compatibility with previous literature and computational frameworks. Using cross-correlation would produce the same results and would not require flipping the kernels during visualization.} operation.
Sufficient padding with zeros is applied on the boundaries to retain the original input size.
We do not need a bias term because it would be applied globally on $\bm{s}^{(i)}$ which is almost equivalent with learning the baseline of $\bm{x}$.

We then pass $\bm{s}^{(i)}$ and a sparsity parameter $d^{(i)}$ in the sparse activation function $\phi$ resulting in the activation map $\bm{\alpha}^{(i)}$:
\begin{equation}
	\label{eq:extrema}
	\bm{\alpha}^{(i)} = \phi(\bm{s}^{(i)}, d^{(i)})
\end{equation}

\noindent
, where $\bm{\alpha}^{(i)}$ is a sparse matrix that its non-zero elements denote the spatial positions of the instances of the $i^{th}$ kernel.
The exact form of $\phi$ and $d^{(i)}$ depends on the choice of the sparse activation function which are presented in Section~\ref{sec:safs}.

We convolve each $\bm{\alpha}^{(i)}$ with its corresponding $\bm{w}^{(i)}$ resulting in the set of $\bm{r}^{(i)}$ which are partial reconstructions of the input:
\begin{equation}
	\label{eq:reconstructions}
	\bm{r}^{(i)} = \bm{\alpha}^{(i)} * \bm{w}^{(i)}
\end{equation}

\noindent
, consisting of sparsely reoccurring patterns of $\bm{w}^{(i)}$ with varying amplitude.
Finally we can reconstruct the input as the sum of the partial reconstructions $\bm{r}^{(i)}$ as follows:
\begin{equation}
	\label{eq:output1}
	\hat{\bm{x}} = \sum\limits_{i=1}^q \bm{r}^{(i)}
\end{equation}

The Mean Absolute Error (MAE) of the input $\bm{x}$ and the prediction $\hat{\bm{x}}$ is then calculated:
\begin{equation}
	\label{eq:lossfunction}
	\mathcal{L}\left( {\bm{x},\hat{\bm{x}}} \right) = \frac{1}{n}\sum\limits_{t=1}^n \Big\lVert\hat{\bm{x}}_t - \bm{x}_t \Big\rVert
\end{equation}

\noindent
, where the index $t$ denotes the $t^{th}$ sample.
The choice of MAE is based on the need to handle outliers in the data with the same weight as normal values.
However, SANs are not restricted in using MAE and other loss functions could be used, such as Mean Square Error.

Using backpropagation, the gradients of the loss $\mathcal{L}$ w.r.t the $\bm{w}^{(i)}$ are calculated:
\begin{equation}
	\label{eq:backpropagation1}
	\nabla\mathcal{L} = \left( \frac{\partial\mathcal{L}}{\partial\bm{w}^{(1)}},\ldots,\frac{\partial\mathcal{L}}{\partial\bm{w}^{(q)}}\right)
\end{equation}

Lastly the $\bm{w}^{(i)}$ are updated using the following learning rule:
\begin{equation}
	\label{eq:backpropagation2}
	\Delta\bm{w}^{(i)} \leftarrow -\lambda\frac{\partial\mathcal{L}}{\partial\bm{w}^{(i)}}
\end{equation}

\noindent
, where $\lambda$ is the learning rate.

After training, we consider $\bm{\alpha}^{(i)}$ (which is calculated during the feed-forward pass from Eq.~\ref{eq:extrema}) and $\bm{w}^{(i)}$ (which is calculated using backpropagation from Eq.~\ref{eq:backpropagation2}) the compressed representation of $\bm{x}$, which can be reconstructed using Equations~\ref{eq:reconstructions} and~\ref{eq:output1}:
\begin{equation}
	\label{eq:output2}
	\hat{\bm{x}} = \sum\limits_{i=1}^q \left(\bm{\alpha}^{(i)} * \bm{w}^{(i)}\right)
\end{equation}

Regarding the $\varphi$ metric and considering Eq.~\ref{eq:output2} our target is to estimate an as accurate as possible representation of $\bm{x}$ through $\bm{\alpha}^{(i)}$ and $\bm{w}^{(i)}$ with the least amount of number of non-zero activations and weights.

The general training procedure of SANs for multiple epochs using batches (instead of one example as previously shown) is presented in Algorithm~\ref{alg:training}.

\begin{algorithm}[H]
	\caption{Sparsely Activated Networks training}\label{alg:training}
	\begin{algorithmic}[1]
		\renewcommand{\algorithmicrequire}{\textbf{Input:}}
		\renewcommand{\algorithmicensure}{\textbf{Output:}}
		\REQUIRE{$\bm{x}$}
		\ENSURE{$\bm{\alpha}$, $\bm{w}$}
		\\ \textit{Hyperparameters:  $m$, $\mu$, $\sigma$, $q$, $\phi$, $d$, $\lambda$, $epochs$, $batches$}
		\FOR{$i$ = 1 to $q$}
		\STATE{$\bm{w}^{(i)} \sim \mathcal{N}(\mu, \sigma)$}
		\ENDFOR{}
		\FOR{$e$ = 1 to $epochs$}
		\FOR{$b$ = 1 to $batches$}
		\STATE{$\bm{x}^{(b)} \sim \bm{x}$}
		\FOR{$i$ = 1 to $q$}
		\STATE{$\bm{s}^{(i)} \leftarrow \bm{x}^{(b)} * \bm{w}^{(i)}$}
		\STATE{$\bm{\alpha}^{(i)} \leftarrow \phi(\bm{s}^{(i)}, d^{(i)})$}
		\STATE{$r^{(i)} \leftarrow \bm{\alpha}^{(i)} * \bm{w}^{(i)}$}
		\ENDFOR{}
		\STATE{$\hat{\bm{x}}^{(b)} \leftarrow \sum\limits_{i=1}^q r^{(i)}$}
		\STATE{$\mathcal{L} \leftarrow \frac{1}{n}\sum\limits_{t=1}^n \lvert\hat{\bm{x}}^{(b)}_t - \bm{x}^{(b)}_t \rvert$}
		\STATE{$\nabla\mathcal{L} = \left( \frac{\partial\mathcal{L}}{\partial\bm{w}^{(1)}},\ldots\frac{\partial\mathcal{L}}{\partial\bm{w}^{(q)}}\right)$}
		\STATE{$\Delta\bm{w}^{(i)} \leftarrow -\lambda\frac{\partial\mathcal{L}}{\partial\bm{w}^{(i)}}$}
		\ENDFOR{}
		\ENDFOR{}
		\RETURN{$\bm{\alpha}$, $\bm{w}$}
	\end{algorithmic}
\end{algorithm}

\section{Experiments}\label{sec:experiments}
For all experiments the weights of the SAN kernels are initialized using the normal distribution $\mathcal{N}(\mu, \sigma)$ with $\mu=0$ and $\sigma=0.1$.
We used Adam~\cite{kingma2014adam} as the optimizer with learning rate $\lambda=0.01$, $b_1=0.9$, $b_2=0.999$, epsilon $\epsilon=10^{-8}$ without weight decay.
For implementing and training SANs we used Pytorch~\cite{paszke2017automatic} with a NVIDIA Titan X Pascal GPU 12GB RAM and a 12 Core Intel i7--8700 CPU @ 3.20GHz on a Linux operating system.

\subsection{Comparison of $\bar\varphi$ for sparse activation functions and various kernel sizes in Physionet}
We study the effect on $\bar\varphi$, of the combined choice of the kernel size $m$ and the sparse activation functions that were defined in Section~\ref{sec:safs}.

\subsubsection{Datasets}
We use one signal from each of 15 signal datasets from Physionet listed in the first column of Table~\ref{table:crrl}.
Each signal consists of $12000$ samples which in turn is split in $12$ signals of $1000$ samples each, to create the training ($6$ signals), validation ($2$ signals) and test datasets ($4$ signals).
The only preprocessing that is done is mean subtraction and division of one standard deviation on the $1000$ samples signals.

\subsubsection{Experiment Setup}
We train five SANs (one for each sparse activation function) for each of the $15$ Physionet datasets, for $30$ epochs with a batch size of $2$ with one kernel of varying size in the range $[1, 250]$.
During validation we selected the models with the kernel size that achieved the best $\bar\varphi$ out of all epochs.
During testing we feed the test data into the selected model and calculate $CR^{-1}$, $\tilde{\mathcal{L}}$ and $\bar\varphi$ for this set of hyperparameters as shown in Table~\ref{table:crrl}.
For Extrema activation we set a `border tolerance' of three samples.

\subsubsection{Results}
\begin{figure*}[!t]
	\includegraphics[width=0.245\textwidth]{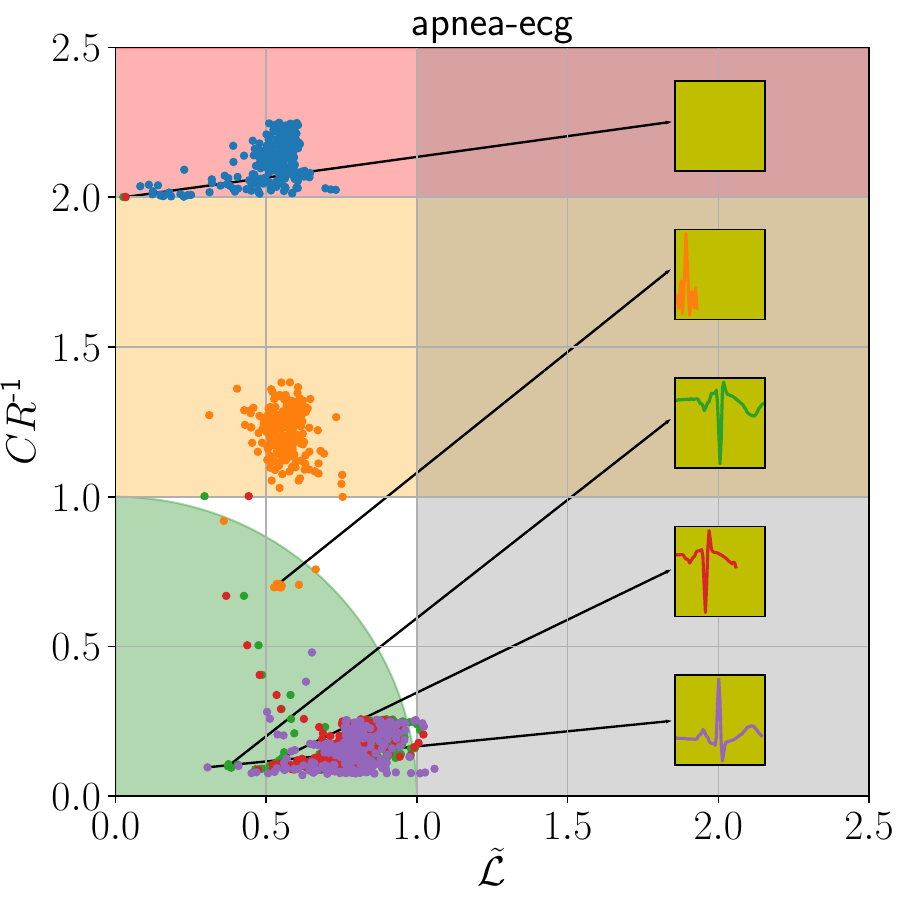}
	\includegraphics[width=0.245\textwidth]{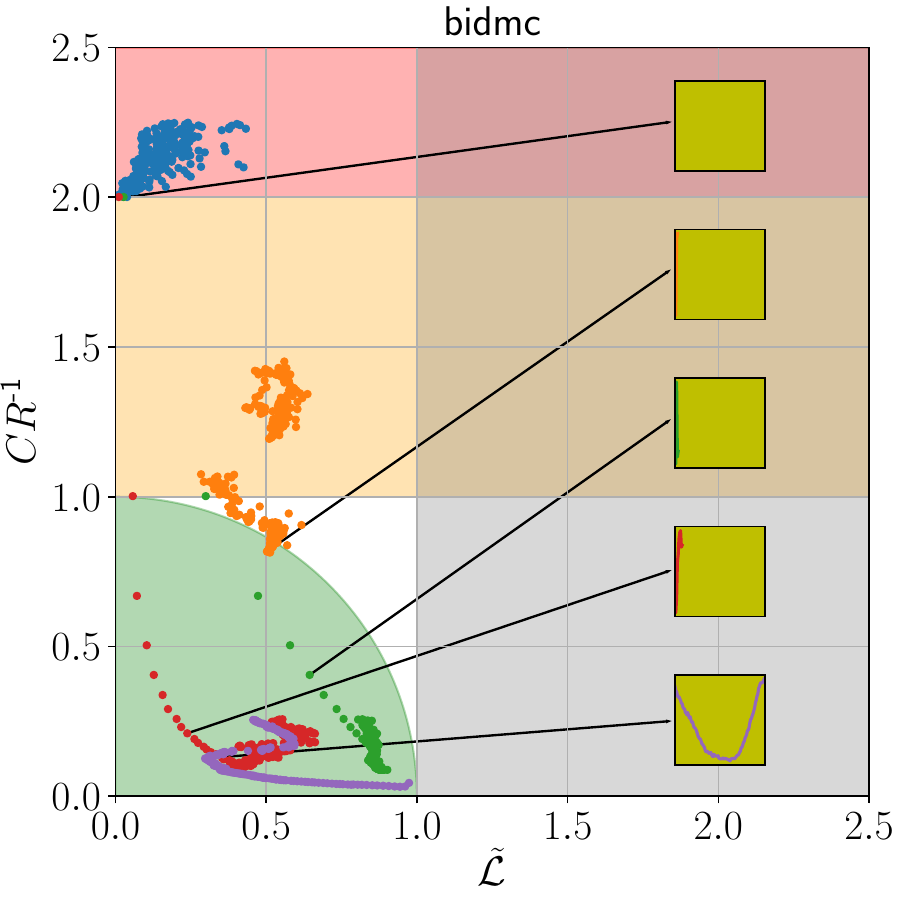}
	\includegraphics[width=0.245\textwidth]{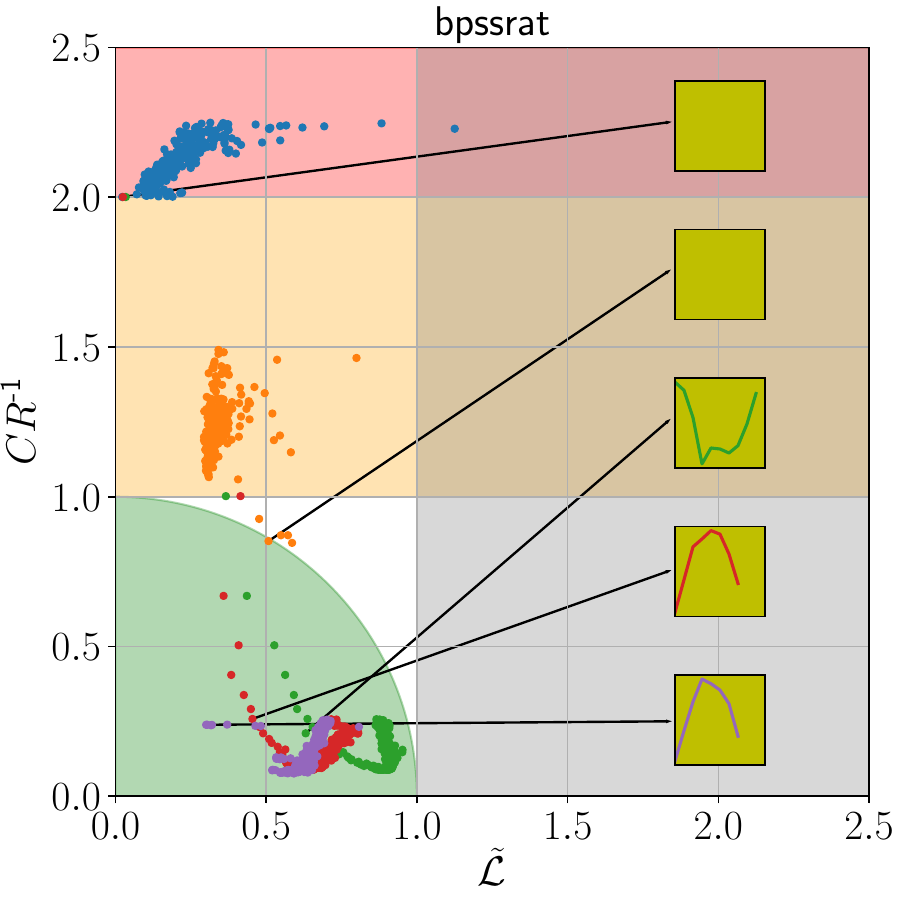}
	\includegraphics[width=0.245\textwidth]{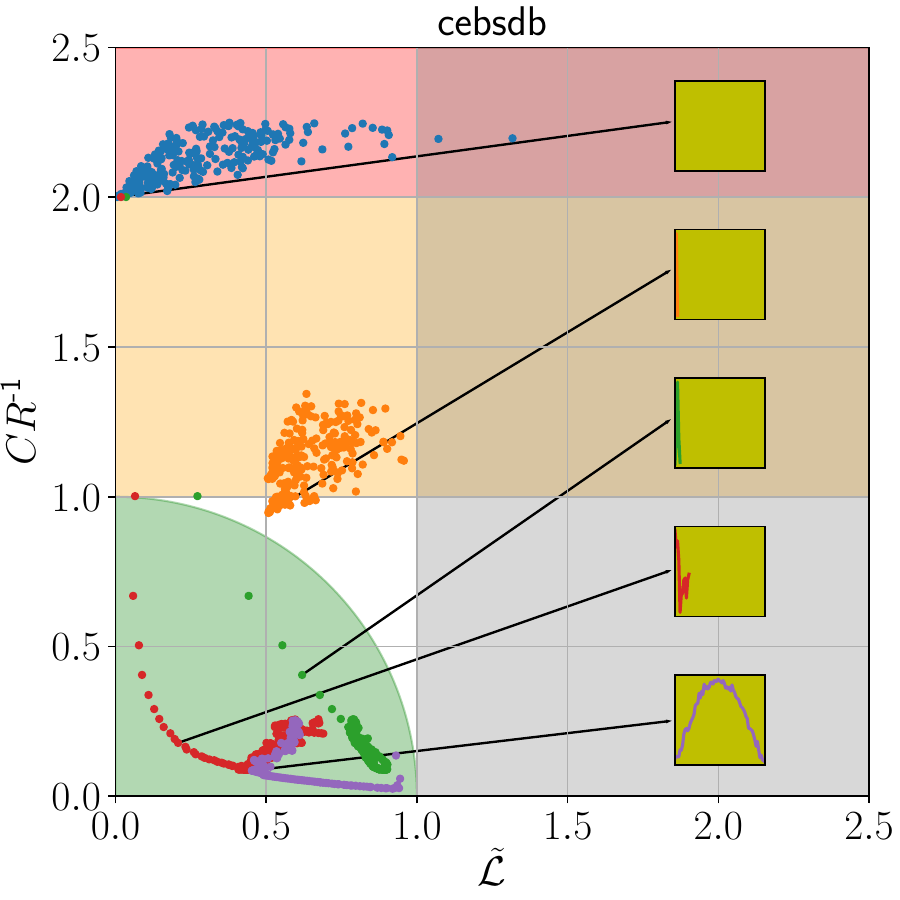}
	\\
	\includegraphics[width=0.245\textwidth]{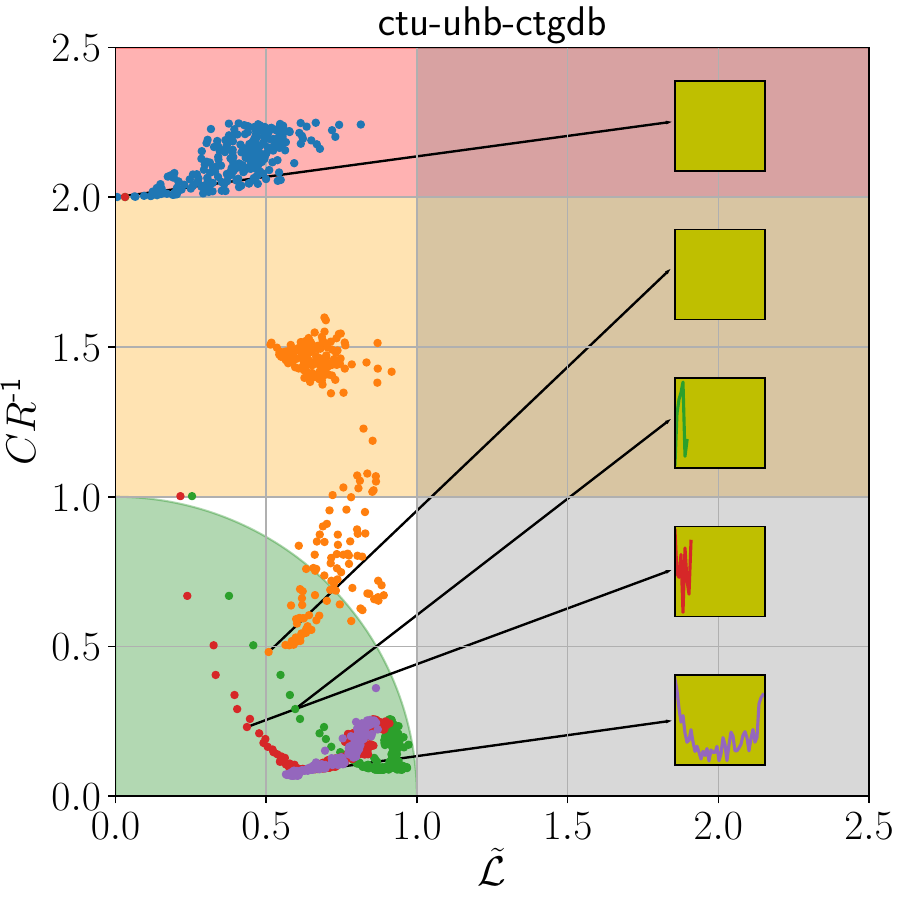}
	\includegraphics[width=0.245\textwidth]{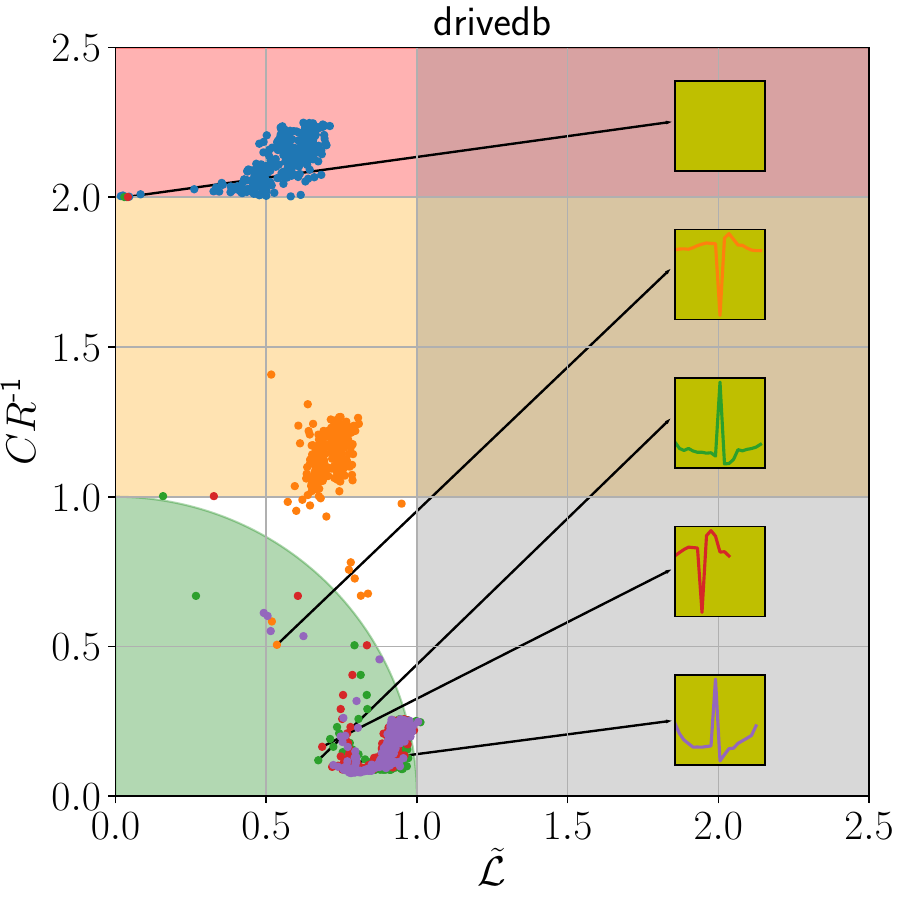}
	\includegraphics[width=0.245\textwidth]{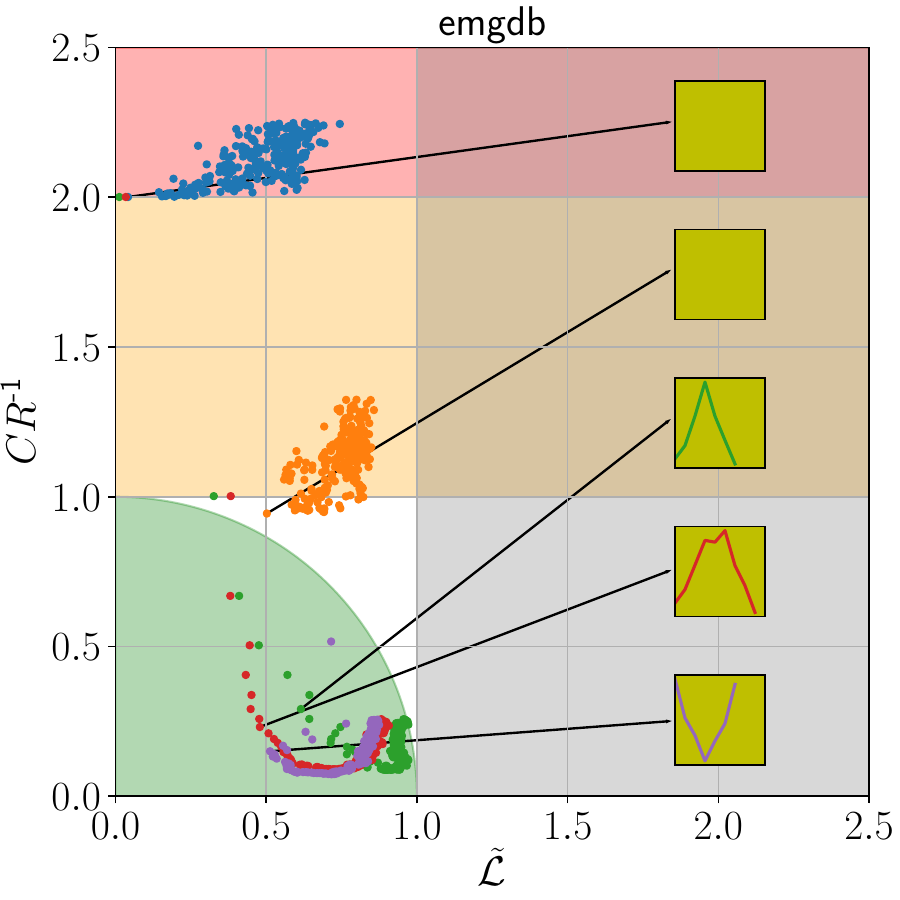}
	\includegraphics[width=0.245\textwidth]{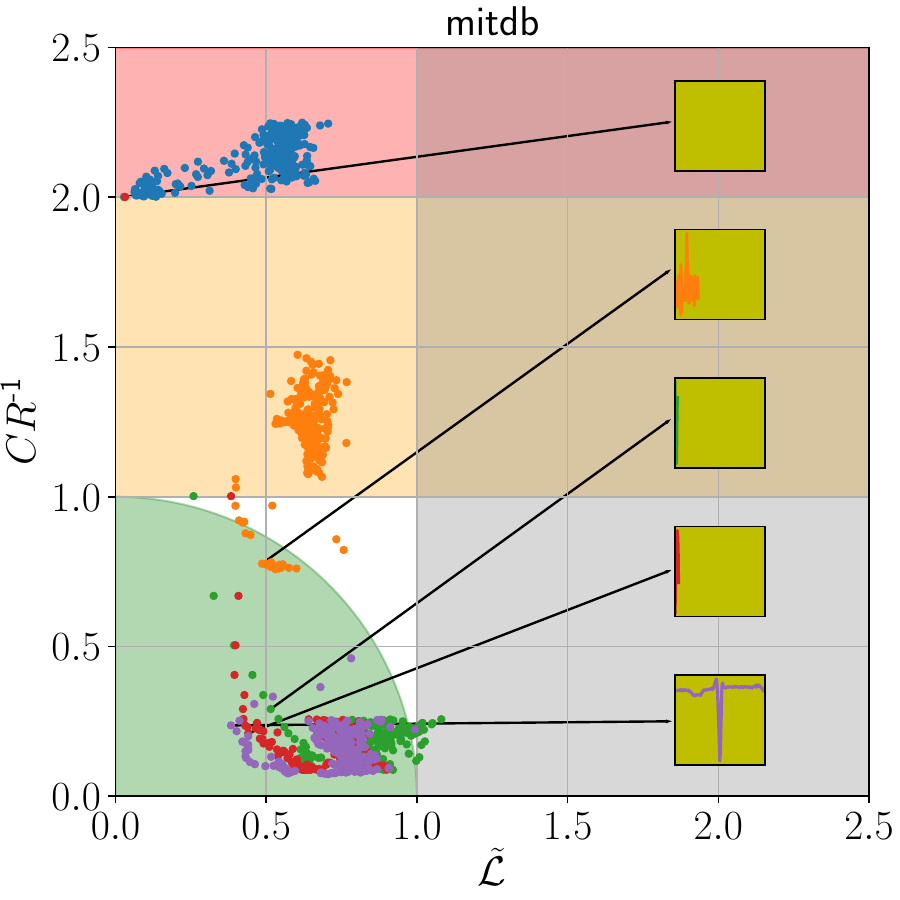}
	\\
	\includegraphics[width=0.245\textwidth]{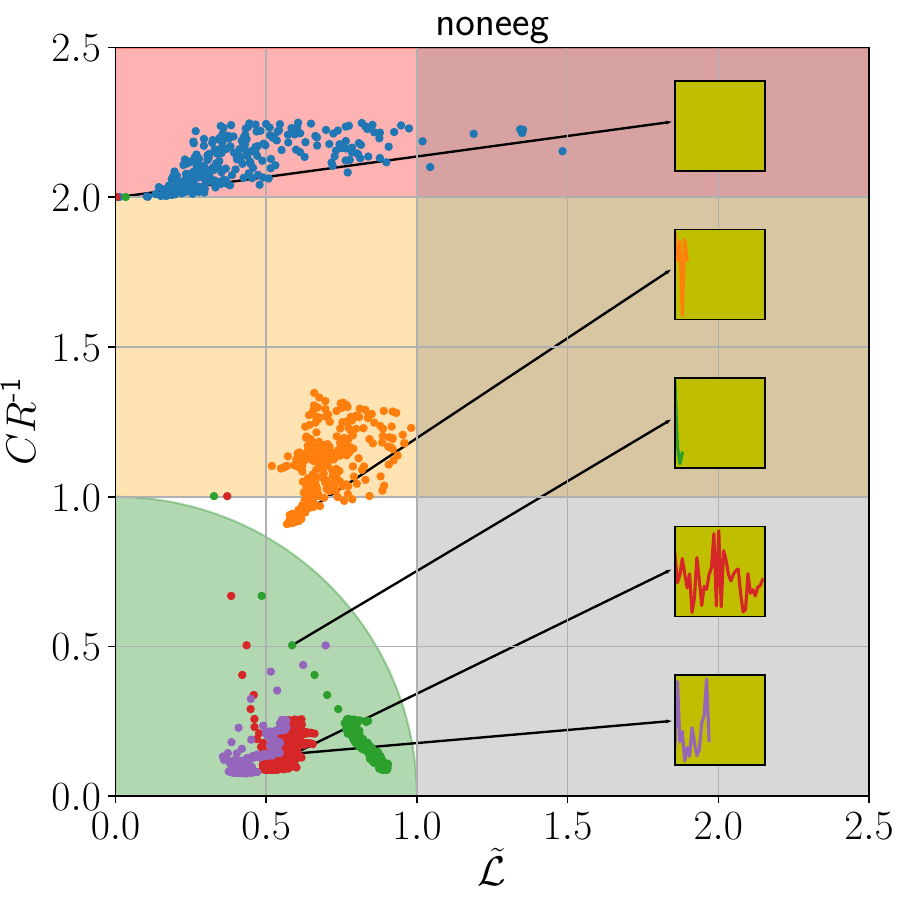}
	\includegraphics[width=0.245\textwidth]{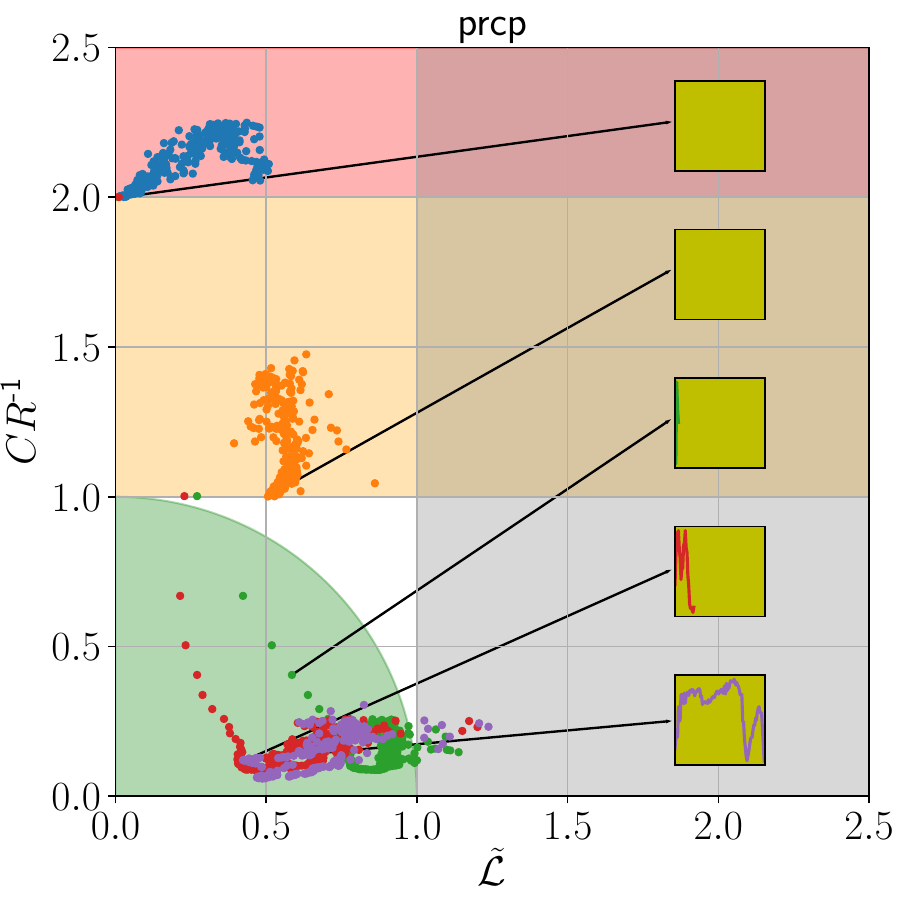}
	\includegraphics[width=0.245\textwidth]{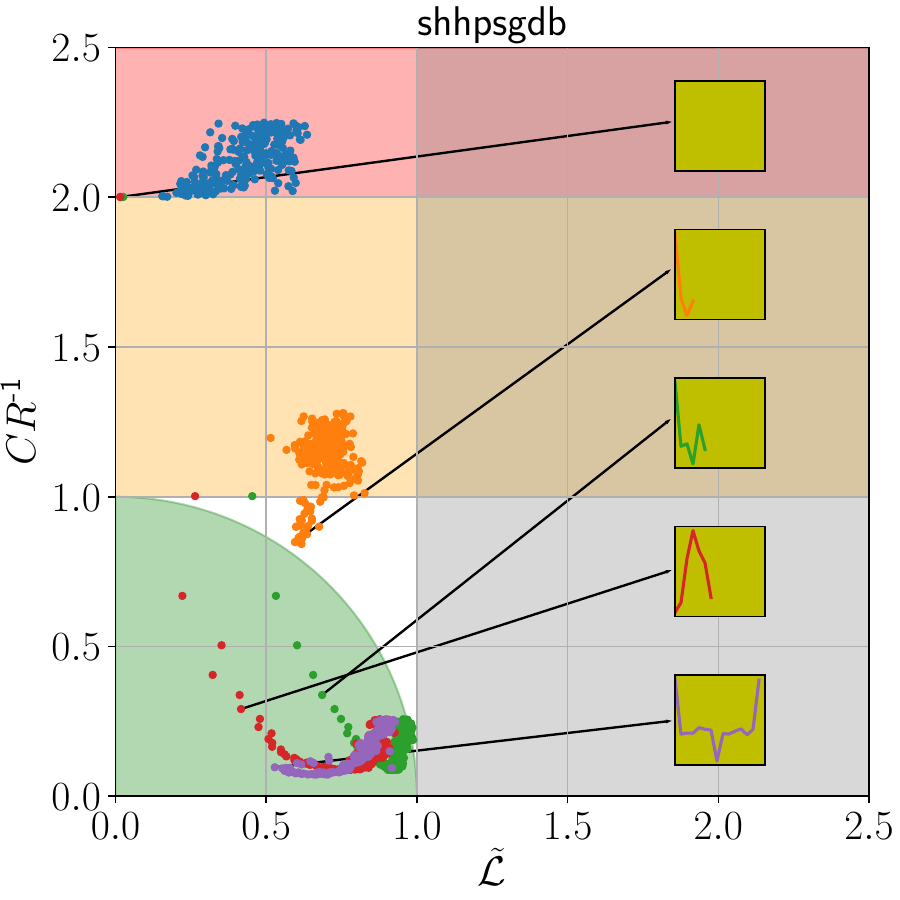}
	\includegraphics[width=0.245\textwidth]{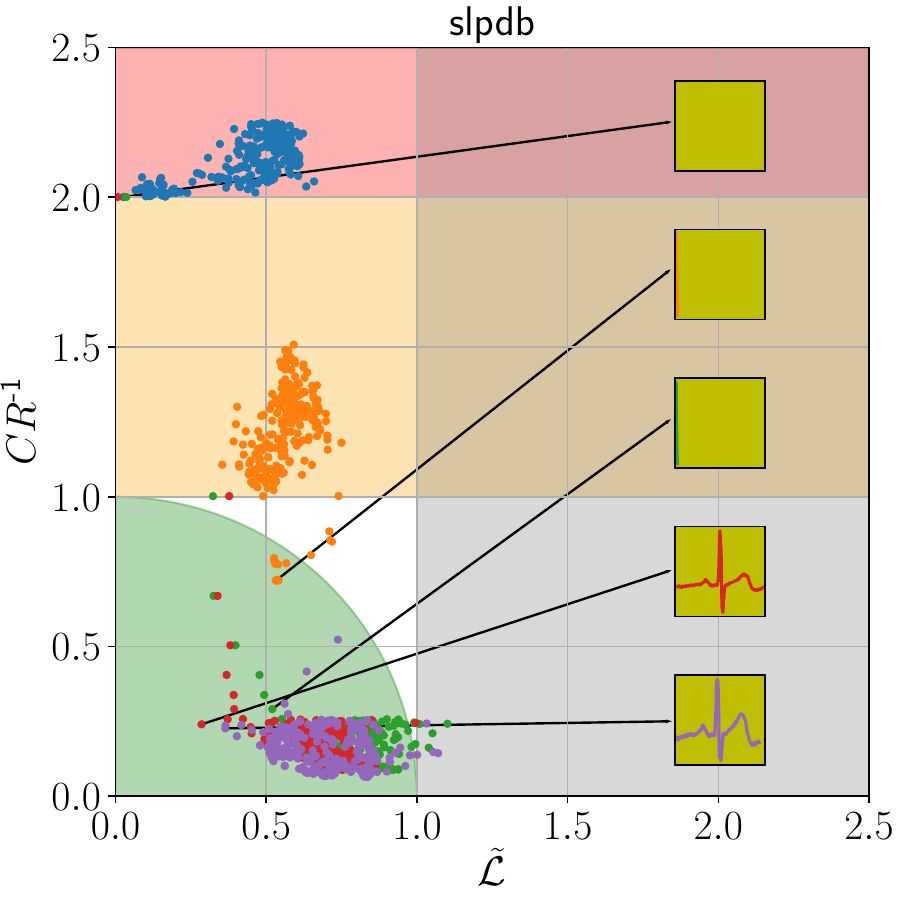}
	\\
	\includegraphics[width=0.245\textwidth]{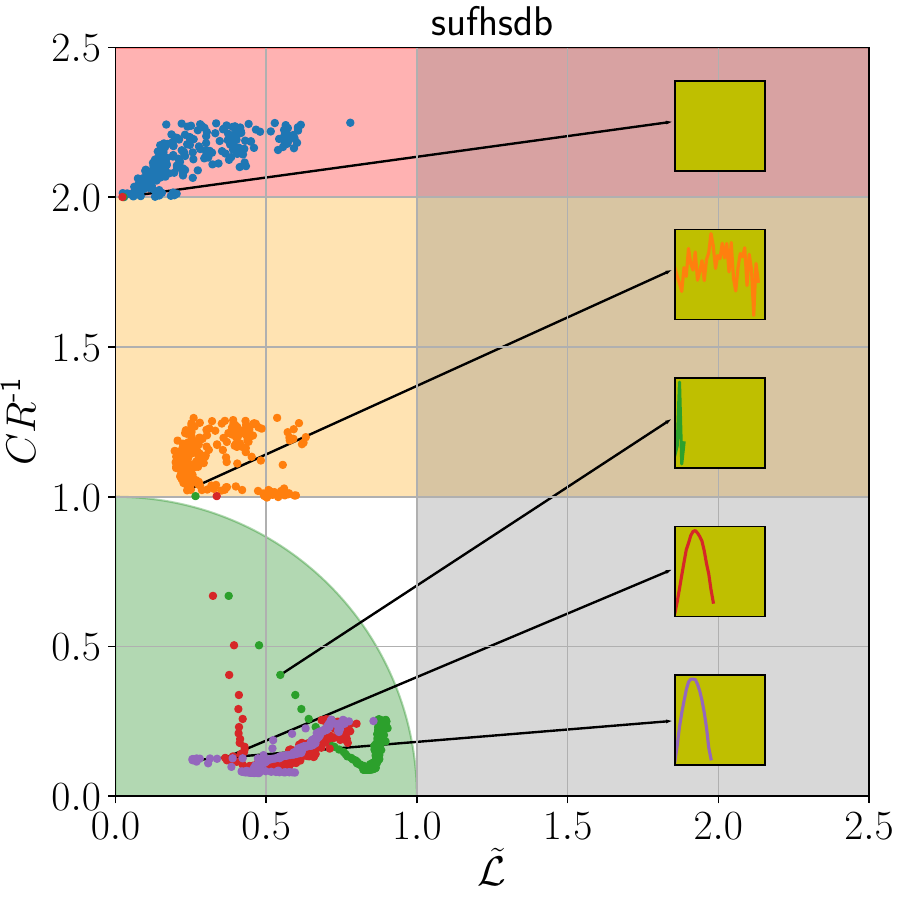}
	\includegraphics[width=0.245\textwidth]{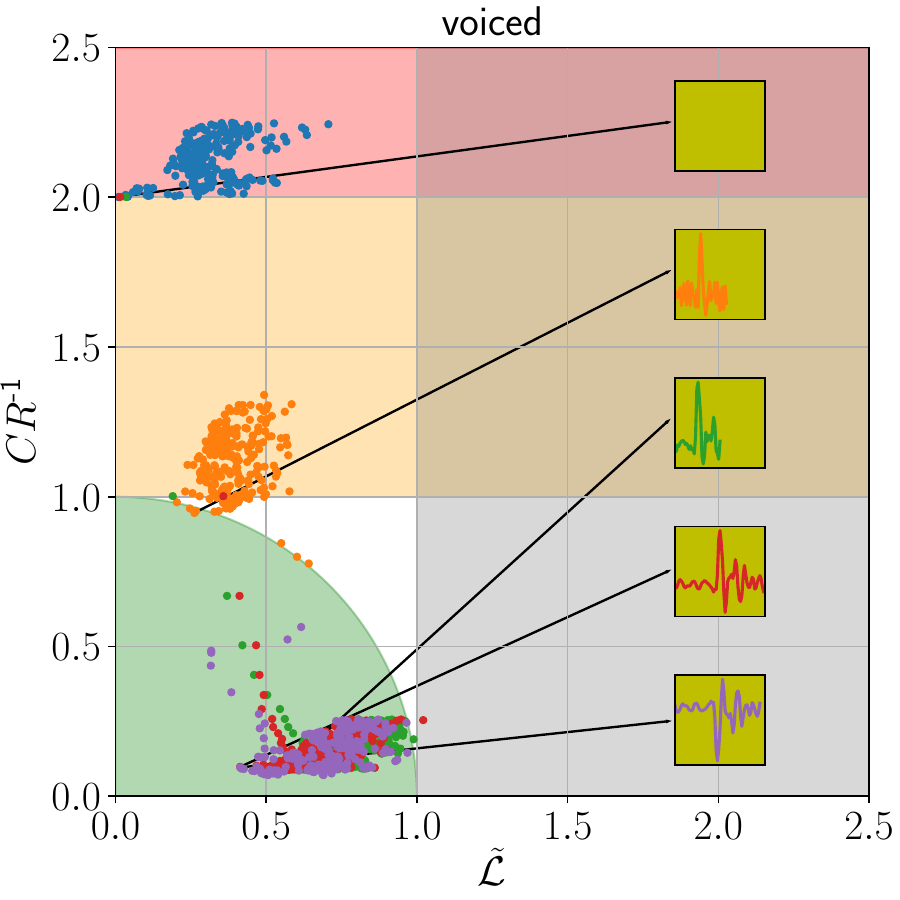}
	\includegraphics[width=0.245\textwidth]{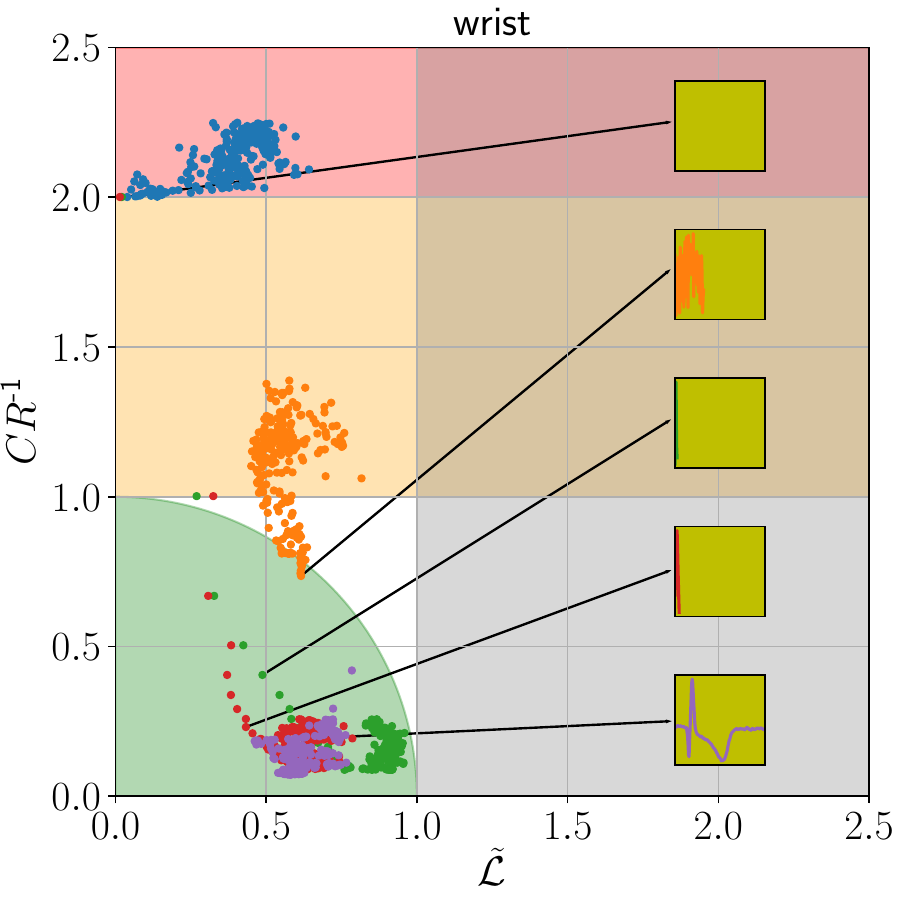}
	\includegraphics[width=0.245\textwidth]{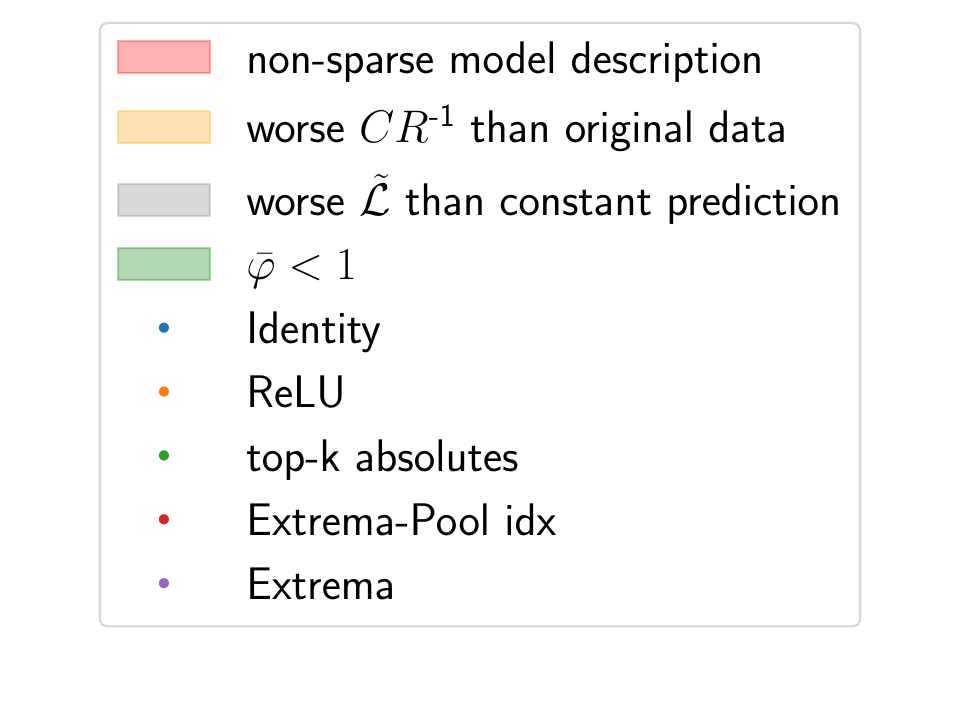}
	\caption{Inverse compression ratio ($CR^{-1}$) vs.\ normalized reconstruction loss ($\tilde{\mathcal{L}}$) for the $15$ datasets of Physionet for various kernel sizes.
	The five inner plots with the yellow background on the right of each subplot, depict the corresponding kernel for the kernel size that achieved the best $\bar\varphi$.}\label{fig:crrl}
\end{figure*}

\begin{figure*}[!t]
	\centering
	\subfloat[Density plot $CR^{-1}$ vs. $\tilde{\mathcal{L}}$]{\includegraphics[width=0.32\textwidth]{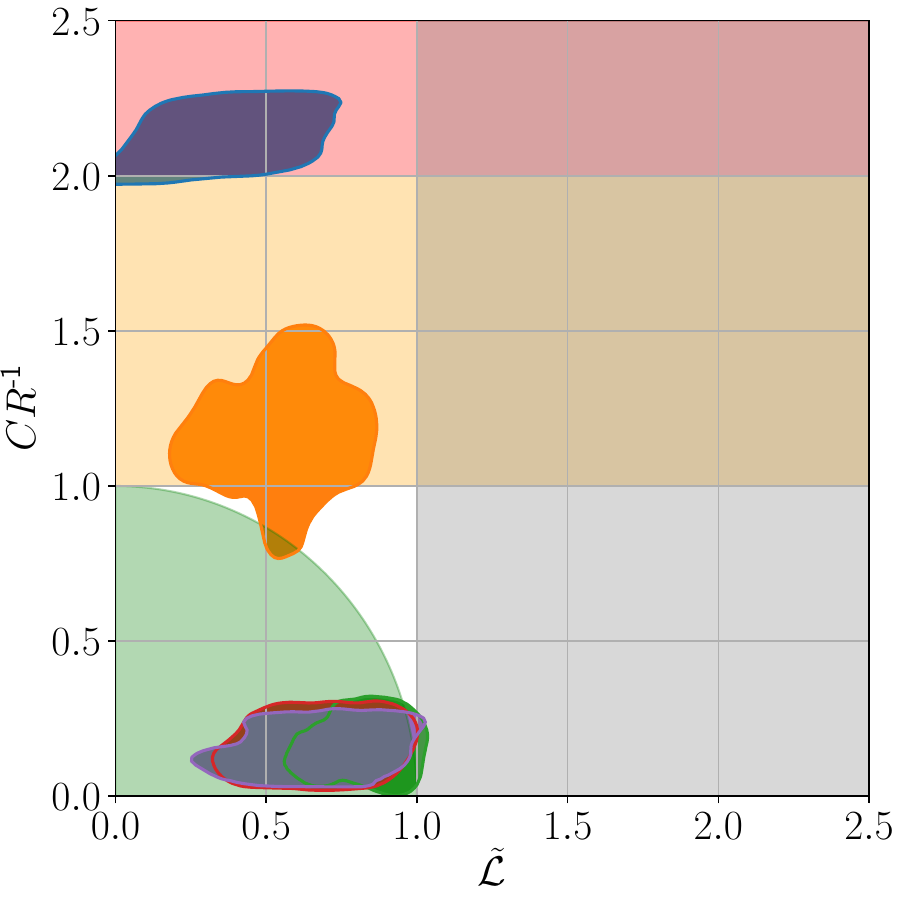}\label{subfig:crrl_density_plot}}
	\subfloat[Confidence intervals $\bar\varphi$ vs. $epochs$]{\includegraphics[width=0.32\textwidth]{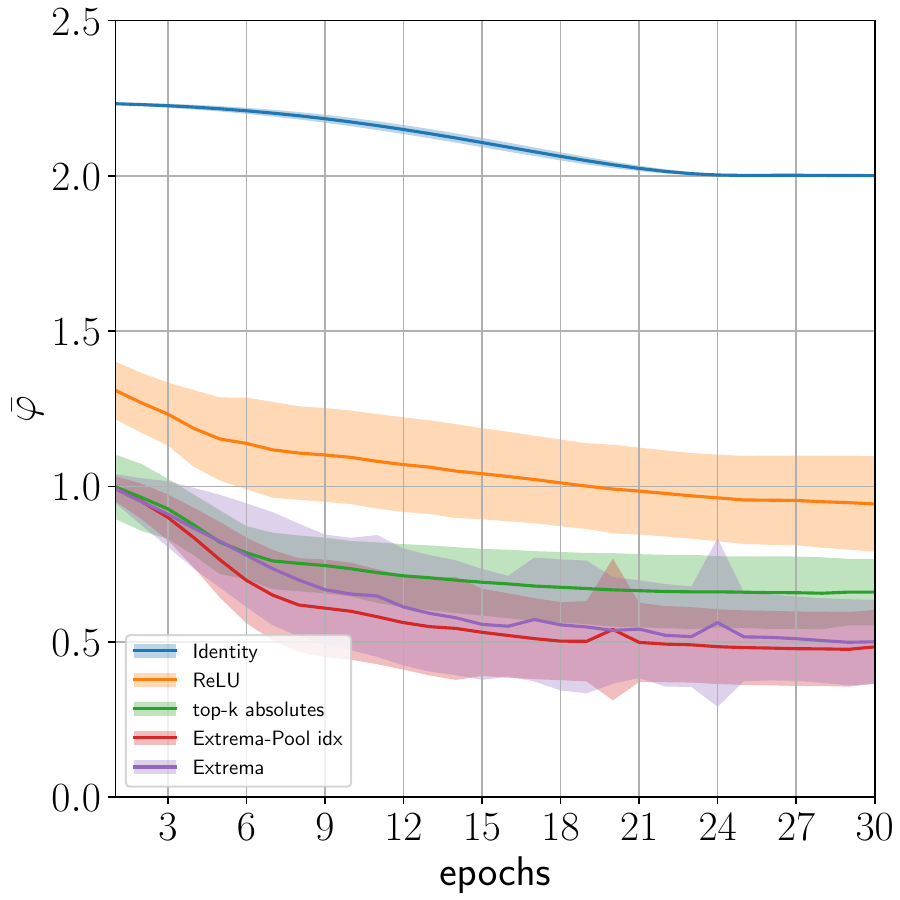}\label{subfig:flithos-epochs}}
	\subfloat[Confidence intervals $\bar\varphi$ vs. $m$]{\includegraphics[width=0.32\textwidth]{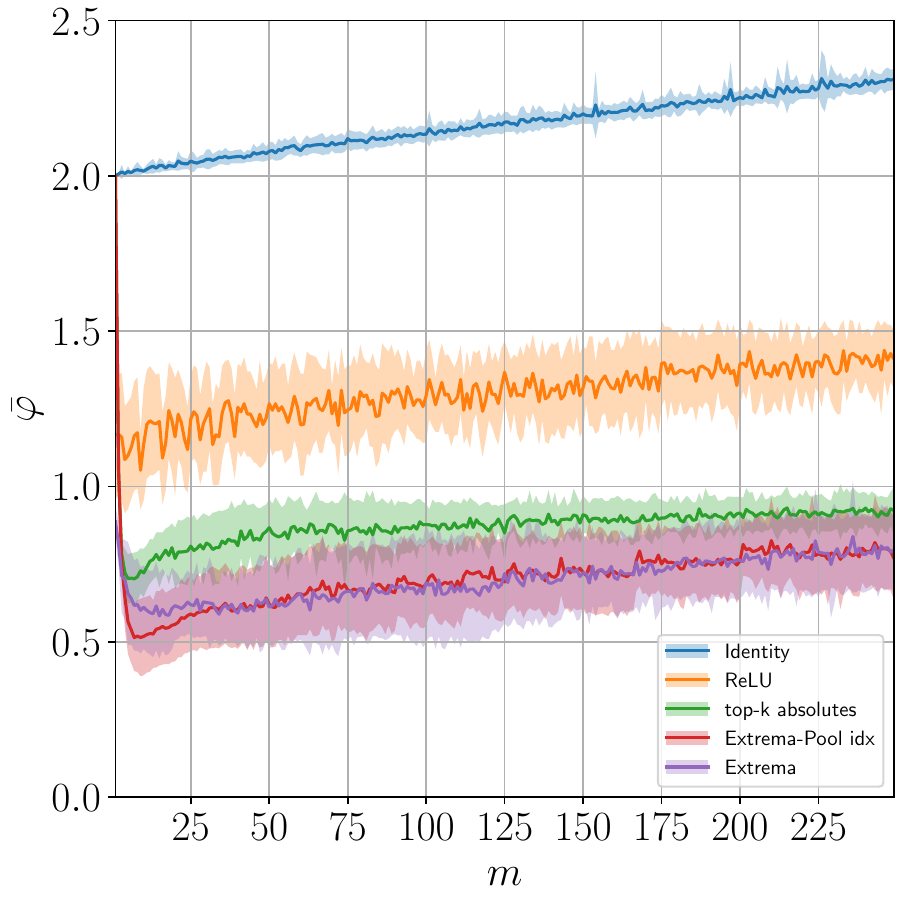}\label{subfig:flithos_m}}
	\caption{Aggregated results of the evaluation of the Physionet databases using the $\bar\varphi$ metric.
	The density plot was created using kernel density estimation with Gaussian kernels and the confidence intervals denote one standard deviation.}\label{fig:flithos}
\end{figure*}

\begin{figure*}[!t]
	\centering
	\includegraphics[width=0.06\textwidth]{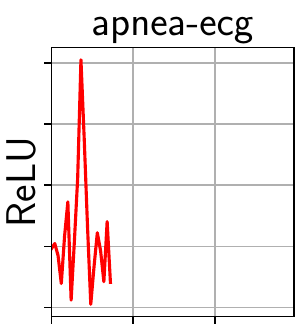}
	\includegraphics[width=0.06\textwidth]{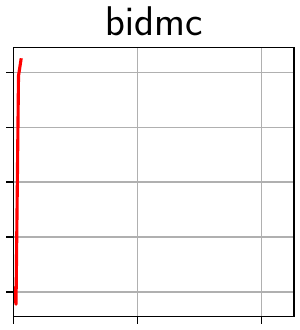}
	\includegraphics[width=0.06\textwidth]{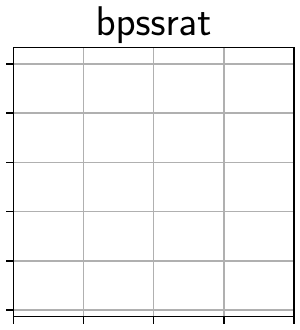}
	\includegraphics[width=0.06\textwidth]{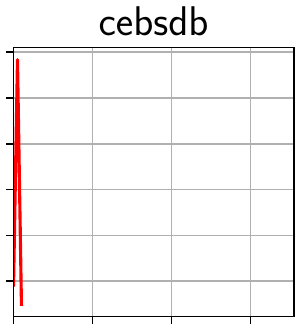}
	\includegraphics[width=0.06\textwidth]{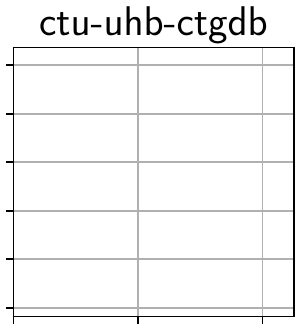}
	\includegraphics[width=0.06\textwidth]{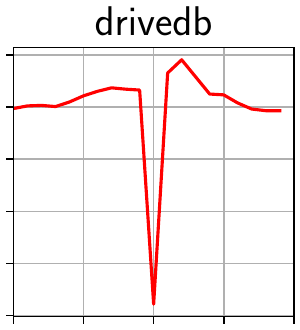}
	\includegraphics[width=0.06\textwidth]{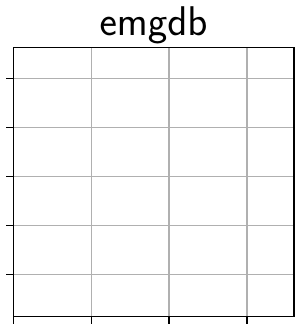}
	\includegraphics[width=0.06\textwidth]{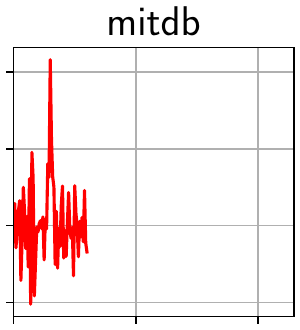}
	\includegraphics[width=0.06\textwidth]{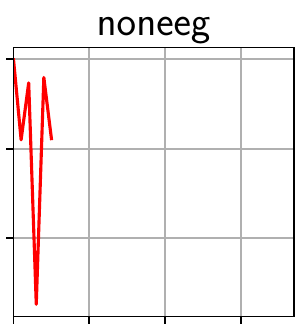}
	\includegraphics[width=0.06\textwidth]{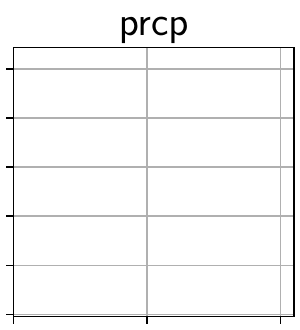}
	\includegraphics[width=0.06\textwidth]{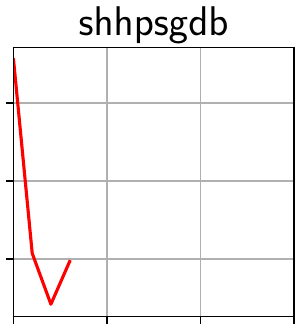}
	\includegraphics[width=0.06\textwidth]{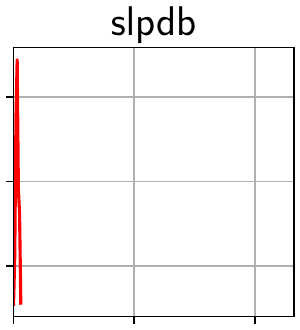}
	\includegraphics[width=0.06\textwidth]{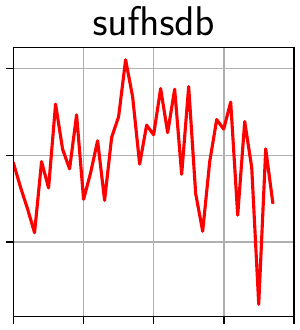}
	\includegraphics[width=0.06\textwidth]{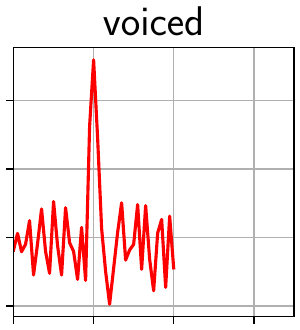}
	\includegraphics[width=0.06\textwidth]{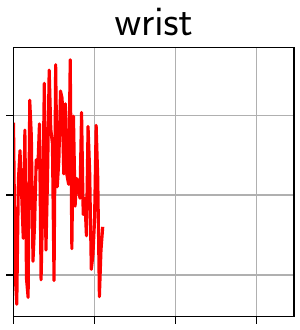}
	\\
	\includegraphics[width=0.06\textwidth]{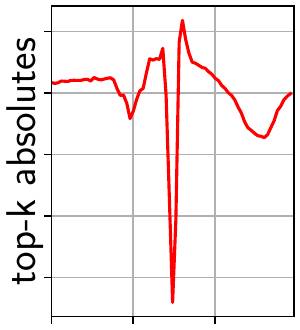}
	\includegraphics[width=0.06\textwidth]{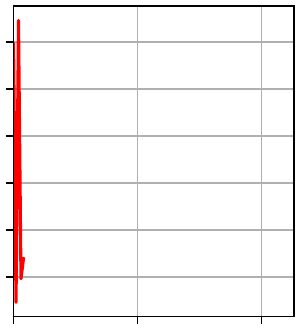}
	\includegraphics[width=0.06\textwidth]{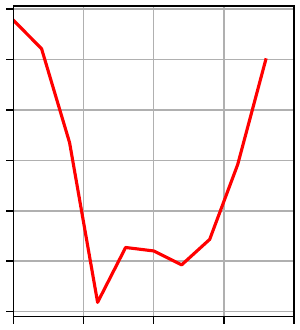}
	\includegraphics[width=0.06\textwidth]{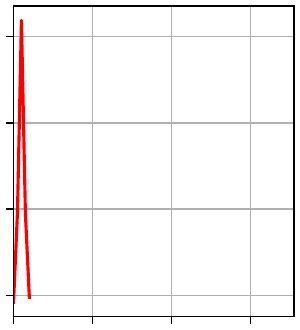}
	\includegraphics[width=0.06\textwidth]{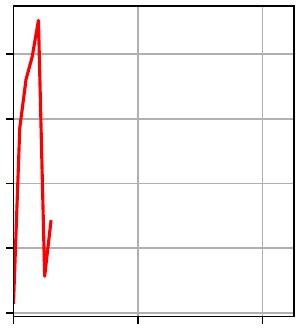}
	\includegraphics[width=0.06\textwidth]{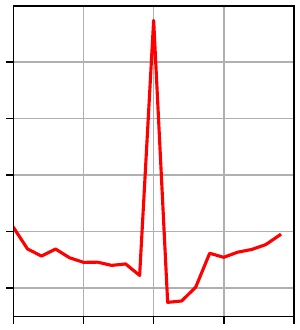}
	\includegraphics[width=0.06\textwidth]{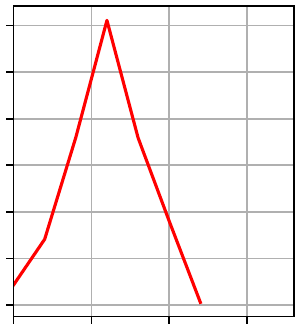}
	\includegraphics[width=0.06\textwidth]{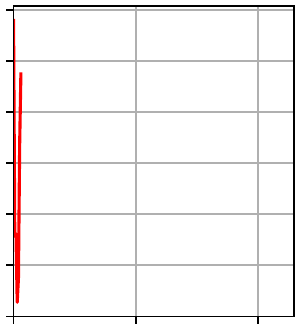}
	\includegraphics[width=0.06\textwidth]{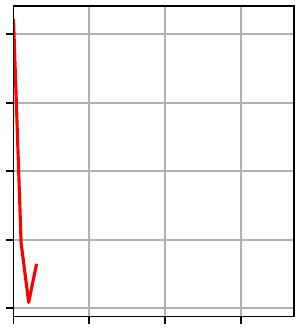}
	\includegraphics[width=0.06\textwidth]{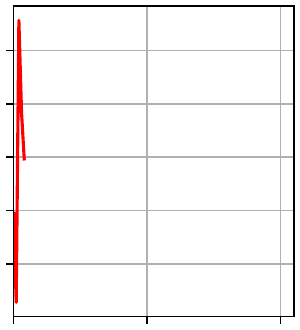}
	\includegraphics[width=0.06\textwidth]{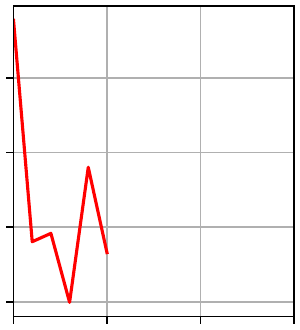}
	\includegraphics[width=0.06\textwidth]{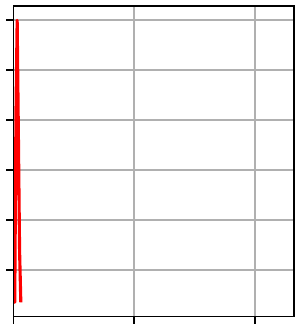}
	\includegraphics[width=0.06\textwidth]{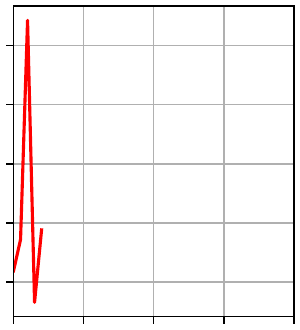}
	\includegraphics[width=0.06\textwidth]{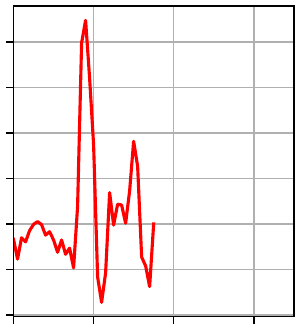}
	\includegraphics[width=0.06\textwidth]{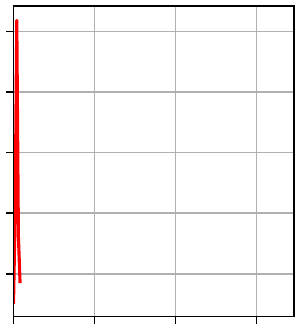}
	\\
	\includegraphics[width=0.06\textwidth]{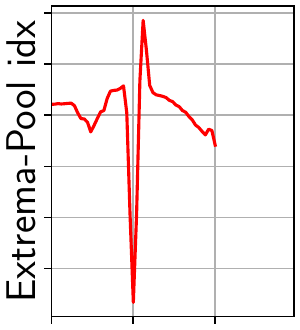}
	\includegraphics[width=0.06\textwidth]{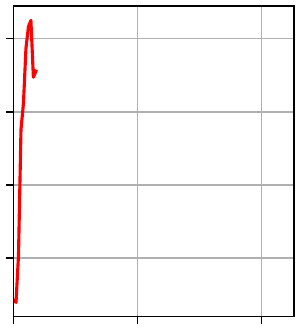}
	\includegraphics[width=0.06\textwidth]{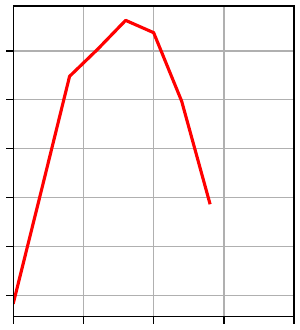}
	\includegraphics[width=0.06\textwidth]{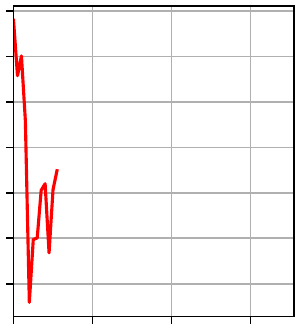}
	\includegraphics[width=0.06\textwidth]{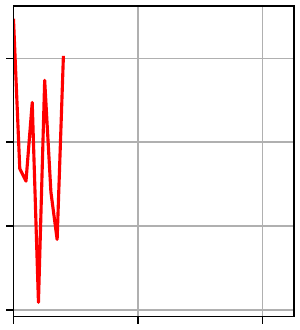}
	\includegraphics[width=0.06\textwidth]{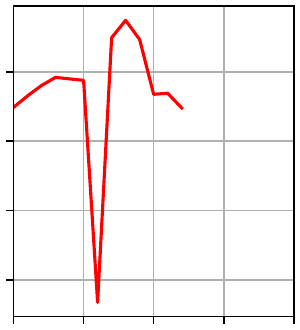}
	\includegraphics[width=0.06\textwidth]{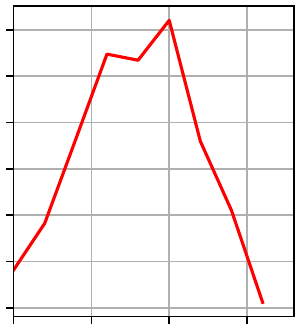}
	\includegraphics[width=0.06\textwidth]{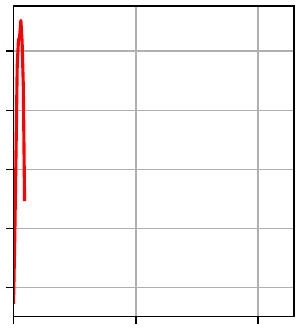}
	\includegraphics[width=0.06\textwidth]{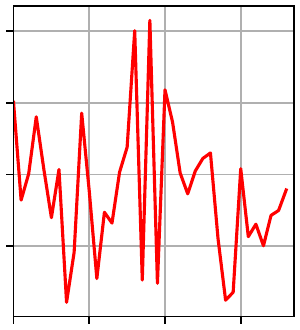}
	\includegraphics[width=0.06\textwidth]{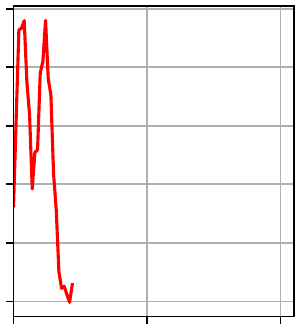}
	\includegraphics[width=0.06\textwidth]{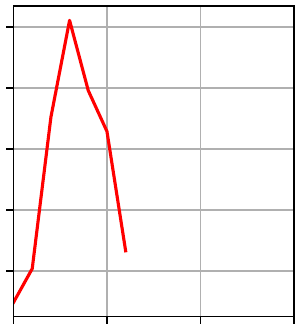}
	\includegraphics[width=0.06\textwidth]{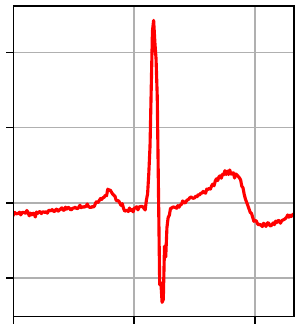}
	\includegraphics[width=0.06\textwidth]{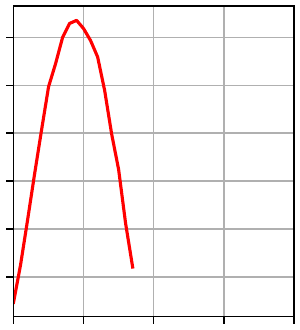}
	\includegraphics[width=0.06\textwidth]{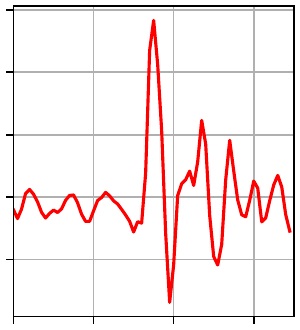}
	\includegraphics[width=0.06\textwidth]{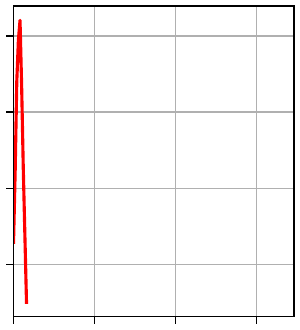}
	\\
	\includegraphics[width=0.06\textwidth]{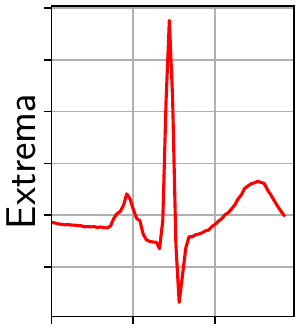}
	\includegraphics[width=0.06\textwidth]{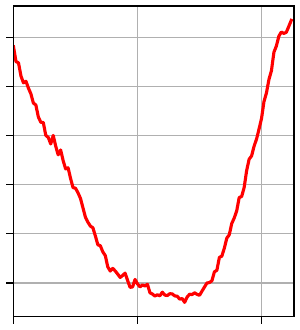}
	\includegraphics[width=0.06\textwidth]{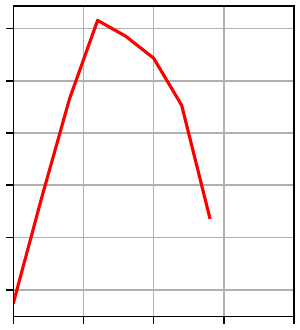}
	\includegraphics[width=0.06\textwidth]{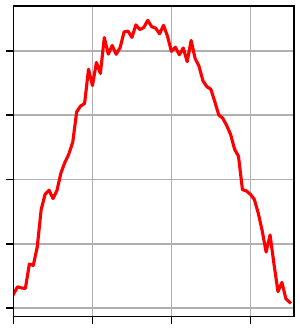}
	\includegraphics[width=0.06\textwidth]{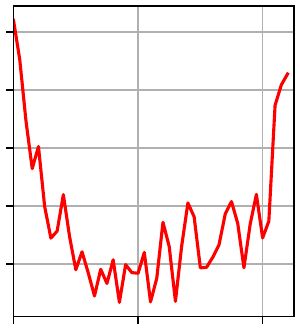}
	\includegraphics[width=0.06\textwidth]{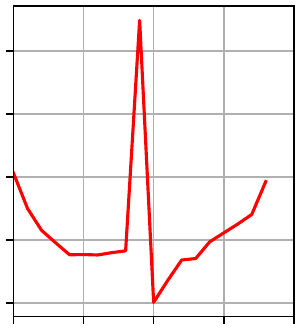}
	\includegraphics[width=0.06\textwidth]{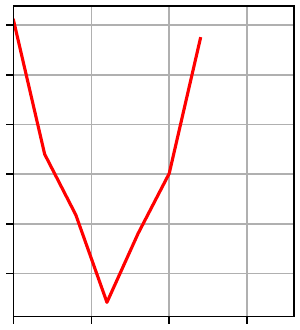}
	\includegraphics[width=0.06\textwidth]{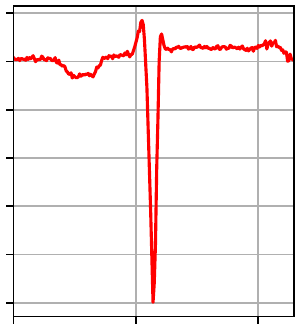}
	\includegraphics[width=0.06\textwidth]{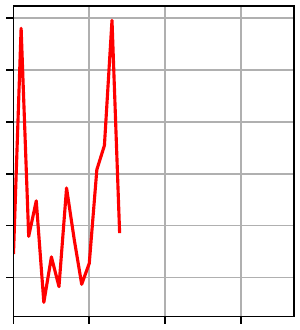}
	\includegraphics[width=0.06\textwidth]{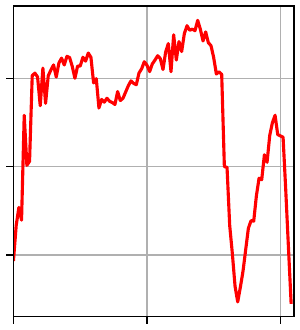}
	\includegraphics[width=0.06\textwidth]{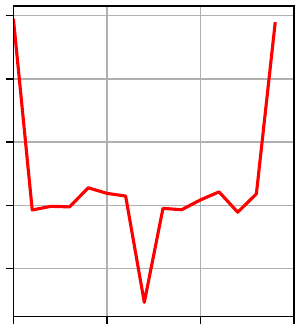}
	\includegraphics[width=0.06\textwidth]{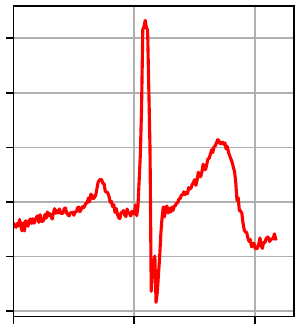}
	\includegraphics[width=0.06\textwidth]{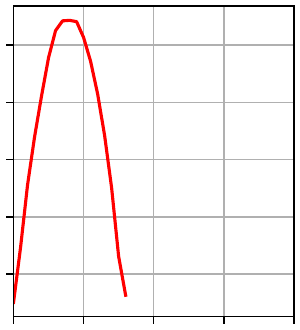}
	\includegraphics[width=0.06\textwidth]{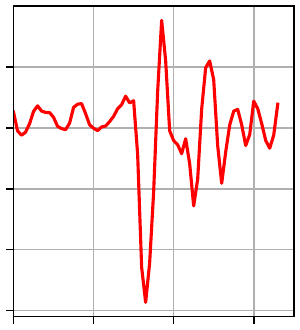}
	\includegraphics[width=0.06\textwidth]{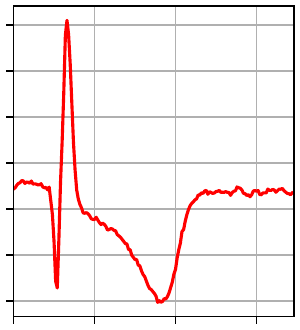}
	\caption{Visualization of the learned kernels for each sparse activation function (row) and for each Physionet database (column).
	}\label{fig:kernelvisualization}
\end{figure*}

The three separate clusters which are depicted in Fig.~\ref{fig:crrl} and the aggregated density plot in Fig.~\ref{fig:flithos}\subref{subfig:crrl_density_plot} between the Identity activation function, the ReLU and the rest show the effect of a sparser activation function on the representation.
The sparser an activation function is the more it compresses, sometimes at the expense of reconstruction error.
However, by visual inspection of Fig.~\ref{fig:kernelvisualization} one could confirm that the learned kernels of the SAN with sparser activation maps (Extrema-Pool indices and Extrema) correspond to the reoccurring patterns in the datasets, thus having high interpretability.
These results suggest that reconstruction error by itself is not a sufficient metric for decomposing data in interpretable components.
Trying to solely achieve lower reconstruction error (such as the case for the Identity activation function) produces noisy learned kernels, while using the combined measure of reconstruction error and compression ratio (smaller $\bar\varphi$) results in interpretable kernels.
Comparing the differences of $\bar\varphi$ between the Identity, the ReLU and the rest sparse activation functions in Fig.~\ref{fig:flithos}\subref{subfig:flithos_m} we notice that the latter produce a minimum region in which we observe interpretable kernels.

\begin{table*}[!t]
	\centering
	\caption{Kernel sizes $m$ with the best $\bar\varphi$ for each sparse activations on Physionet databases}\label{table:crrl}
	\begin{adjustbox}{width=\textwidth}
		\begin{tabular}{lrrrrrrrrrrrrrrrrrrrr}
\toprule
{} & \multicolumn{4}{c}{Identity} & \multicolumn{4}{c}{ReLU} & \multicolumn{4}{c}{top-k absolutes} & \multicolumn{4}{c}{Extrema-Pool idx} & \multicolumn{4}{c}{Extrema} \\
{} &      $m$ & $CR\textsuperscript{-1}$ & $\tilde{\mathcal{L}}$ & $\bar\varphi$ &  $m$ & $CR\textsuperscript{-1}$ & $\tilde{\mathcal{L}}$ & $\bar\varphi$ &             $m$ & $CR\textsuperscript{-1}$ & $\tilde{\mathcal{L}}$ & $\bar\varphi$ &              $m$ & $CR\textsuperscript{-1}$ & $\tilde{\mathcal{L}}$ & $\bar\varphi$ &     $m$ & $CR\textsuperscript{-1}$ & $\tilde{\mathcal{L}}$ & $\bar\varphi$ \\
\textbf{Datasets     } &          &                          &                       &               &      &                          &                       &               &                 &                          &                       &               &                  &                          &                       &               &         &                          &                       &               \\
\midrule
\textbf{apnea-ecg    } &        1 &                     2.00 &                  0.03 &          2.00 &   19 &                     0.70 &                  0.53 &          0.87 &              74 &                     0.10 &                  0.37 &          0.39 &               51 &                     0.09 &                  0.47 &          0.48 &      72 &                     0.10 &                  0.31 &          0.32 \\
\textbf{bidmc        } &        1 &                     2.00 &                  0.04 &          2.00 &    4 &                     0.82 &                  0.50 &          0.96 &               5 &                     0.41 &                  0.64 &          0.76 &               10 &                     0.21 &                  0.24 &          0.32 &     113 &                     0.13 &                  0.30 &          0.32 \\
\textbf{bpssrat      } &        1 &                     2.00 &                  0.02 &          2.00 &    1 &                     0.85 &                  0.51 &          0.99 &              10 &                     0.21 &                  0.63 &          0.67 &                8 &                     0.26 &                  0.45 &          0.52 &       8 &                     0.24 &                  0.30 &          0.38 \\
\textbf{cebsdb       } &        1 &                     2.00 &                  0.01 &          2.00 &    3 &                     0.95 &                  0.51 &          1.07 &               5 &                     0.41 &                  0.62 &          0.74 &               12 &                     0.18 &                  0.21 &          0.28 &      71 &                     0.09 &                  0.45 &          0.46 \\
\textbf{ctu-uhb-ctgdb} &        1 &                     2.00 &                  0.01 &          2.00 &    1 &                     0.48 &                  0.51 &          0.71 &               7 &                     0.29 &                  0.60 &          0.66 &                9 &                     0.23 &                  0.44 &          0.49 &      45 &                     0.07 &                  0.57 &          0.57 \\
\textbf{drivedb      } &        1 &                     2.00 &                  0.04 &          2.00 &   20 &                     0.51 &                  0.54 &          0.74 &              20 &                     0.12 &                  0.67 &          0.68 &               13 &                     0.17 &                  0.69 &          0.71 &      19 &                     0.10 &                  0.72 &          0.73 \\
\textbf{emgdb        } &        1 &                     2.00 &                  0.04 &          2.00 &    1 &                     0.94 &                  0.50 &          1.07 &               7 &                     0.29 &                  0.62 &          0.68 &                9 &                     0.23 &                  0.48 &          0.53 &       7 &                     0.15 &                  0.51 &          0.53 \\
\textbf{mitdb        } &        1 &                     2.00 &                  0.03 &          2.00 &   61 &                     0.78 &                  0.49 &          0.92 &               7 &                     0.29 &                  0.52 &          0.59 &               10 &                     0.21 &                  0.44 &          0.49 &     229 &                     0.24 &                  0.38 &          0.45 \\
\textbf{noneeg       } &        1 &                     2.00 &                  0.01 &          2.00 &    6 &                     0.91 &                  0.57 &          1.08 &               4 &                     0.50 &                  0.59 &          0.77 &               37 &                     0.09 &                  0.49 &          0.50 &      15 &                     0.12 &                  0.36 &          0.38 \\
\textbf{prcp         } &        1 &                     2.00 &                  0.03 &          2.00 &    1 &                     1.00 &                  0.51 &          1.12 &               5 &                     0.41 &                  0.59 &          0.71 &               23 &                     0.11 &                  0.41 &          0.42 &     105 &                     0.12 &                  0.42 &          0.44 \\
\textbf{shhpsgdb     } &        1 &                     2.00 &                  0.02 &          2.00 &    4 &                     0.85 &                  0.60 &          1.05 &               6 &                     0.34 &                  0.69 &          0.77 &                7 &                     0.29 &                  0.42 &          0.51 &      15 &                     0.10 &                  0.53 &          0.54 \\
\textbf{slpdb        } &        1 &                     2.00 &                  0.03 &          2.00 &    7 &                     0.72 &                  0.53 &          0.90 &               7 &                     0.29 &                  0.52 &          0.60 &              232 &                     0.24 &                  0.29 &          0.37 &     218 &                     0.23 &                  0.36 &          0.43 \\
\textbf{sufhsdb      } &        1 &                     2.00 &                  0.03 &          2.00 &   38 &                     1.02 &                  0.24 &          1.05 &               5 &                     0.41 &                  0.55 &          0.68 &               18 &                     0.13 &                  0.36 &          0.39 &      17 &                     0.12 &                  0.26 &          0.28 \\
\textbf{voiced       } &        1 &                     2.00 &                  0.01 &          2.00 &   41 &                     0.95 &                  0.26 &          0.98 &              36 &                     0.09 &                  0.56 &          0.57 &               70 &                     0.10 &                  0.41 &          0.43 &      67 &                     0.10 &                  0.41 &          0.43 \\
\textbf{wrist        } &        1 &                     2.00 &                  0.04 &          2.00 &   56 &                     0.74 &                  0.62 &          0.96 &               5 &                     0.41 &                  0.49 &          0.63 &                9 &                     0.23 &                  0.43 &          0.49 &     173 &                     0.18 &                  0.46 &          0.50 \\
\bottomrule
\end{tabular}

	\end{adjustbox}
\end{table*}

\subsection{Evaluation of the reconstruction of SANs using a Supervised CNN on UCI-epilepsy}
We study the quality of the reconstructions of SANs by training a supervised 1D Convolutional Neural Network (CNN) on the output of each SAN\@.
We also study the effect that $m$ has on $\bar\varphi$ and the accuracy of the classifier for the five sparse activation functions.

\subsubsection{Dataset}
We use the UCI-epilepsy recognition dataset that consists of $500$ signals each one with $4097$ samples (23.5 seconds).
The dataset is annotated into five classes with $100$ signals for each class.
For the purposes of this paper we use a variation of the database\footnote{\url{https://archive.ics.uci.edu/ml/datasets/Epileptic+Seizure+Recognition}} in which the EEG signals are split into segments with $178$ samples each, resulting in a balanced dataset that consists of $11500$ EEG signals in total.

\subsubsection{Experiment Setup}
First, we merge the tumor classes ($2$ and $3$) and the eyes classes ($4$ and $5$) resulting in a modified dataset of three classes (tumor, eyes, epilepsy).
We then split the $11500$ signals into $76\%$, $12\%$ and $12\%$ ($8740,1380,1380$ signals) as training, validation and test data respectively and normalize in the range $[0, 1]$ using the global max and min.
For the SANs, we used two kernels $q=2$ with a varying size in the range of $[8, 15]$ and trained for $5$ epochs with a batch size of $64$.
After training, we choose the model that performed the lowest $\bar\varphi$ out of all epochs.

During supervised learning the weights of the kernels are frozen and a CNN is stacked on top of the reconstruction output of the SANs.
The CNN feature extractor consists of two convolutional layers with $3$ and $16$ filters and kernel size $5$, each one followed by a ReLU and a Max-Pool with pool size $2$.
The classifier consists of three fully connected layers with $656$, $120$ and $84$ units.
The first two fully connected layers are followed by a ReLU while the last one produces the predictions.
The CNN is trained for an additional $5$ epochs with the same batch size and model selection procedure as with SANs and categorical cross-entropy as the loss function.
For Extrema activation we set a `border tolerance' of two samples.

\subsubsection{Results}
As shown in Table.~\ref{table:uciepilepsysupervised}, although we use a significantly reduced representation size, the classification accuracy differences (A\textsubscript{$\pm$\%}) are retained which suggests that SANs choose the most important features to represent the data.
For example for $m=12$ for the Extrema activation function, there is an increase in accuracy of $3\%$ (the baseline CNN on the original data achieved $\DTLfetch{keys-values}{key}{uci-epilepsy-supervised-accuracy}{value}\%$) although a reduced representation is used with only $34\%$ of the size w.r.t the original data.

\begin{table*}[!t]
	\centering
	\caption{SANs with supervised stacked CNN for UCI-epilepsy Classification}\label{table:uciepilepsysupervised}
	\begin{adjustbox}{width=\textwidth}
		\begin{tabular}{lrrrrrrrrrrrrrrrrrrrr}
\toprule
{} & \multicolumn{4}{c}{Identity} & \multicolumn{4}{c}{ReLU} & \multicolumn{4}{c}{top-k absolutes} & \multicolumn{4}{c}{Extrema-Pool idx} & \multicolumn{4}{c}{Extrema} \\
{} & $CR\textsuperscript{-1}$ & $\tilde{\mathcal{L}}$ & $\bar\varphi$ & A\textsubscript{$\pm$\%} & $CR\textsuperscript{-1}$ & $\tilde{\mathcal{L}}$ & $\bar\varphi$ & A\textsubscript{$\pm$\%} & $CR\textsuperscript{-1}$ & $\tilde{\mathcal{L}}$ & $\bar\varphi$ & A\textsubscript{$\pm$\%} & $CR\textsuperscript{-1}$ & $\tilde{\mathcal{L}}$ & $\bar\varphi$ & A\textsubscript{$\pm$\%} & $CR\textsuperscript{-1}$ & $\tilde{\mathcal{L}}$ & $\bar\varphi$ & A\textsubscript{$\pm$\%} \\
\textbf{$m$} &                          &                       &               &                          &                          &                       &               &                          &                          &                       &               &                          &                          &                       &               &                          &                          &                       &               &                          \\
\midrule
\textbf{8  } &                     4.09 &                  0.04 &          4.09 &                     +5.6 &                     2.09 &                  0.02 &          2.09 &                     +4.3 &                     0.58 &                  0.73 &          0.93 &                     -0.4 &                     0.58 &                  0.41 &          0.71 &                     -7.8 &                     0.37 &                  0.45 &          0.59 &                     +2.8 \\
\textbf{9  } &                     4.10 &                  0.03 &          4.10 &                     +1.4 &                     2.10 &                  0.03 &          2.10 &                    -44.3 &                     0.53 &                  0.74 &          0.91 &                     -1.2 &                     0.53 &                  0.41 &          0.67 &                     -2.6 &                     0.35 &                  0.42 &          0.56 &                     +0.8 \\
\textbf{10 } &                     4.11 &                  0.02 &          4.11 &                     +8.0 &                     3.61 &                  0.11 &          3.62 &                    -26.6 &                     0.49 &                  0.76 &          0.91 &                    -13.4 &                     0.49 &                  0.43 &          0.65 &                     +1.7 &                     0.35 &                  0.42 &          0.56 &                     +2.4 \\
\textbf{11 } &                     4.12 &                  0.02 &          4.12 &                    -44.3 &                     2.12 &                  0.03 &          2.13 &                     +6.2 &                     0.48 &                  0.76 &          0.90 &                    -12.2 &                     0.48 &                  0.41 &          0.63 &                     -1.2 &                     0.34 &                  0.41 &          0.54 &                     +1.0 \\
\textbf{12 } &                     4.13 &                  0.02 &          4.13 &                     +6.6 &                     2.13 &                  0.03 &          2.13 &                     +2.1 &                     0.45 &                  0.77 &          0.89 &                     -5.7 &                     0.45 &                  0.44 &          0.63 &                     -1.9 &                     0.34 &                  0.42 &          0.55 &                     +3.0 \\
\textbf{13 } &                     4.15 &                  0.06 &          4.15 &                     +4.1 &                     2.15 &                  0.03 &          2.15 &                     +6.5 &                     0.44 &                  0.78 &          0.89 &                    -13.5 &                     0.44 &                  0.44 &          0.62 &                     -9.3 &                     0.34 &                  0.42 &          0.55 &                     +0.8 \\
\textbf{14 } &                     4.16 &                  0.06 &          4.16 &                     +6.2 &                     3.66 &                  0.11 &          3.66 &                    -44.3 &                     0.43 &                  0.78 &          0.89 &                    -10.4 &                     0.43 &                  0.46 &          0.63 &                     -2.1 &                     0.34 &                  0.43 &          0.55 &                     -1.4 \\
\textbf{15 } &                     4.17 &                  0.03 &          4.17 &                     +9.1 &                     2.17 &                  0.03 &          2.17 &                     -8.4 &                     0.42 &                  0.78 &          0.89 &                     -3.5 &                     0.42 &                  0.46 &          0.62 &                     +1.8 &                     0.34 &                  0.43 &          0.55 &                     -2.0 \\
\bottomrule
\end{tabular}

	\end{adjustbox}
\end{table*}

\subsection{Evaluation of the reconstruction of SANs using a Supervised FNN on MNIST and FMNIST}
\subsubsection{Dataset}
For the same task as the previous one but for 2D, we use MNIST which consists of a training dataset of $60000$ greyscale images with handwritten digits and a test dataset of $10000$ images each one having size of $28\times 28$.
The exact same procedure is used for FMNIST~\cite{xiao2017fashion}.

\subsubsection{Experiment Setup}
The models consist of two kernels $q=2$ with a varying size in the range of $[1, 6]$.
We use $10000$ images from the training dataset as a validation dataset and train on the rest $50000$ for $5$ epochs with a batch size of $64$.
We do not apply any preprocessing on the images.

During supervised learning the weights of the kernels are frozen and a one layer fully connected network (FNN) is stacked on top of the reconstruction output of the SANs.
The FNN is trained for an additional $5$ epochs with the same batch size and model selection procedure as with SANs and categorical cross-entropy as the loss function.
For Extrema activation we set a `border tolerance' of two samples.

\subsubsection{Results}
As shown in Table.~\ref{table:mnistsupervised} the accuracies achieved by the reconstructions of some SANs do not drop proportionally compared to those of an FNN trained on the original data ($\DTLfetch{keys-values}{key}{mnist-supervised-accuracy}{value}\%$), although they have been heavily compressed.
It is interesting to note that in some cases SANs reconstructions, such as for the Extrema-Pool indices, performed even better than the original data.
This suggests the overwhelming presence of redundant information that resides in the raw pixels of the original data and further indicates that SANs extract the most representative features of the data.

\begin{table*}[!t]
	\centering
	\caption{SAN with supervised stacked FNN on MNIST}\label{table:mnistsupervised}
	\begin{adjustbox}{width=\textwidth}
		\begin{tabular}{lrrrrrrrrrrrrrrrrrrrr}
\toprule
{} & \multicolumn{4}{c}{Identity} & \multicolumn{4}{c}{ReLU} & \multicolumn{4}{c}{top-k absolutes} & \multicolumn{4}{c}{Extrema-Pool idx} & \multicolumn{4}{c}{Extrema} \\
{} & $CR\textsuperscript{-1}$ & $\tilde{\mathcal{L}}$ & $\bar\varphi$ & A\textsubscript{$\pm$\%} & $CR\textsuperscript{-1}$ & $\tilde{\mathcal{L}}$ & $\bar\varphi$ & A\textsubscript{$\pm$\%} & $CR\textsuperscript{-1}$ & $\tilde{\mathcal{L}}$ & $\bar\varphi$ & A\textsubscript{$\pm$\%} & $CR\textsuperscript{-1}$ & $\tilde{\mathcal{L}}$ & $\bar\varphi$ & A\textsubscript{$\pm$\%} & $CR\textsuperscript{-1}$ & $\tilde{\mathcal{L}}$ & $\bar\varphi$ & A\textsubscript{$\pm$\%} \\
\textbf{$m$} &                          &                       &               &                          &                          &                       &               &                          &                          &                       &               &                          &                          &                       &               &                          &                          &                       &               &                          \\
\midrule
\textbf{1  } &                     1.16 &                  0.00 &          1.16 &                     -0.1 &                     1.16 &                  0.01 &          1.16 &                     -1.6 &                     1.16 &                  0.00 &          1.16 &                     -0.3 &                     1.16 &                  0.00 &          1.16 &                     -0.1 &                     0.09 &                  0.89 &          0.90 &                    -11.3 \\
\textbf{2  } &                     1.55 &                  0.01 &          1.55 &                     -0.3 &                     0.78 &                  0.01 &          0.78 &                     -0.0 &                     1.37 &                  0.02 &          1.37 &                     -0.6 &                     0.48 &                  0.61 &          0.79 &                     +1.3 &                     0.09 &                  0.83 &          0.83 &                     -7.6 \\
\textbf{3  } &                     1.93 &                  0.00 &          1.93 &                     -0.5 &                     1.16 &                  0.03 &          1.16 &                     -0.8 &                     0.63 &                  0.25 &          0.68 &                     -1.6 &                     0.30 &                  0.51 &          0.59 &                     +0.0 &                     0.08 &                  0.50 &          0.51 &                     -7.3 \\
\textbf{4  } &                     2.30 &                  0.03 &          2.30 &                     -0.5 &                     1.53 &                  0.02 &          1.53 &                     -0.6 &                     0.39 &                  0.40 &          0.55 &                     -3.7 &                     0.22 &                  0.59 &          0.63 &                     -0.6 &                     0.07 &                  0.56 &          0.57 &                    -10.3 \\
\textbf{5  } &                     2.66 &                  0.09 &          2.66 &                     -0.8 &                     0.73 &                  0.03 &          0.73 &                     -0.2 &                     0.20 &                  0.55 &          0.59 &                     -5.3 &                     0.16 &                  0.60 &          0.62 &                     -0.7 &                     0.07 &                  0.57 &          0.57 &                     -8.8 \\
\textbf{6  } &                     3.02 &                  0.06 &          3.02 &                     -0.6 &                     0.60 &                  0.01 &          0.60 &                     -0.5 &                     0.14 &                  0.61 &          0.63 &                     -8.8 &                     0.12 &                  0.63 &          0.65 &                     -1.7 &                     0.05 &                  0.60 &          0.61 &                    -11.4 \\
\bottomrule
\end{tabular}

	\end{adjustbox}
\end{table*}

\begin{table*}[!t]
	\centering
	\caption{SAN with supervised stacked FNN on FashionMNIST}\label{table:fashionmnistsupervised}
	\begin{adjustbox}{width=\textwidth}
		\begin{tabular}{lrrrrrrrrrrrrrrrrrrrr}
\toprule
{} & \multicolumn{4}{c}{Identity} & \multicolumn{4}{c}{ReLU} & \multicolumn{4}{c}{top-k absolutes} & \multicolumn{4}{c}{Extrema-Pool idx} & \multicolumn{4}{c}{Extrema} \\
{} & $CR\textsuperscript{-1}$ & $\tilde{\mathcal{L}}$ & $\bar\varphi$ & A\textsubscript{$\pm$\%} & $CR\textsuperscript{-1}$ & $\tilde{\mathcal{L}}$ & $\bar\varphi$ & A\textsubscript{$\pm$\%} & $CR\textsuperscript{-1}$ & $\tilde{\mathcal{L}}$ & $\bar\varphi$ & A\textsubscript{$\pm$\%} & $CR\textsuperscript{-1}$ & $\tilde{\mathcal{L}}$ & $\bar\varphi$ & A\textsubscript{$\pm$\%} & $CR\textsuperscript{-1}$ & $\tilde{\mathcal{L}}$ & $\bar\varphi$ & A\textsubscript{$\pm$\%} \\
\textbf{$m$} &                          &                       &               &                          &                          &                       &               &                          &                          &                       &               &                          &                          &                       &               &                          &                          &                       &               &                          \\
\midrule
\textbf{1  } &                     3.00 &                  0.01 &          3.00 &                     -1.7 &                     1.50 &                  0.00 &          1.50 &                     +1.0 &                     3.00 &                  0.01 &          3.00 &                     -0.5 &                     3.00 &                  0.01 &          3.00 &                     -3.8 &                     0.35 &                  0.86 &          0.93 &                     -4.0 \\
\textbf{2  } &                     3.58 &                  0.01 &          3.58 &                     +2.0 &                     3.00 &                  0.05 &          3.00 &                     +1.3 &                     1.50 &                  0.31 &          1.55 &                     -5.0 &                     0.99 &                  0.58 &          1.16 &                     -0.6 &                     0.23 &                  0.81 &          0.84 &                     -6.6 \\
\textbf{3  } &                     3.94 &                  0.05 &          3.94 &                     +2.0 &                     2.01 &                  0.02 &          2.01 &                     +1.0 &                     0.63 &                  0.63 &          0.89 &                     -7.0 &                     0.48 &                  0.52 &          0.72 &                     -0.7 &                     0.16 &                  0.65 &          0.68 &                     -9.3 \\
\textbf{4  } &                     4.22 &                  0.06 &          4.22 &                     -0.1 &                     0.01 &                  1.00 &          1.00 &                    -71.5 &                     0.39 &                  0.69 &          0.79 &                     -9.4 &                     0.32 &                  0.61 &          0.70 &                     -1.9 &                     0.09 &                  0.70 &          0.71 &                     -9.6 \\
\textbf{5  } &                     4.45 &                  0.04 &          4.45 &                     +1.8 &                     1.92 &                  0.02 &          1.92 &                     +1.3 &                     0.20 &                  0.75 &          0.78 &                    -15.0 &                     0.19 &                  0.60 &          0.63 &                     -5.0 &                     0.07 &                  0.64 &          0.64 &                    -12.2 \\
\textbf{6  } &                     4.70 &                  0.04 &          4.70 &                     +0.8 &                     2.80 &                  0.05 &          2.80 &                     +2.2 &                     0.14 &                  0.77 &          0.78 &                    -19.6 &                     0.13 &                  0.66 &          0.67 &                     -4.3 &                     0.05 &                  0.69 &          0.70 &                    -13.7 \\
\bottomrule
\end{tabular}

	\end{adjustbox}
\end{table*}

\section{Discussion}\label{sec:discussion}
SANs combined with the $\varphi$ metric compress the description of the data in a way a minimum description language framework would, by encoding them into $\bm{w}^{(i)}$ and $\bm{\alpha}^{(i)}$.
The experiments done in Section~\ref{sec:experiments} show that the use of Identity, ReLU and (in a lesser degree) top-k absolutes produce noisy features, while on the other hand Extrema-Pool indices and Extrema produce more robust features and can be configured with parameters (kernel size and $med$) with values that can be derived by simple visual inspection of the data.

From the point of view of Sparse Dictionary Learning, SANs kernels could be seen as the atoms of a learned dictionary specializing in interpretable pattern matching (e.g.\ for Electrocardiogram (ECG) input the kernels of SANs are ECG beats) and the sparse activation map as the representation.
The fact that SANs are wide with fewer larger kernel sizes instead of deep with smaller kernel sizes make them more interpretable than the DNNs and in some cases without sacrificing significant accuracy.

An advantage of SANs compared to Sparse Autoencoders~\cite{ng2011sparse} is that the constrain of activation proximity can be applied individually in each example instead of requiring the computation of forward-pass of all examples.
Additionally, SANs create exact zeros instead near-zeros, which reduces co-adaptation between instances of the neuron activation.

$\varphi$ could be seen as an alternative formalization of Occam's razor~\cite{soklakov2002occam} to Solomonov's theory of inductive inference~\cite{solomonoff1964formal} but with a deterministic interpretation instead of a probabilistic one.
The cost of the description of the data could be seen as proportional to the number of weights and the number of non-zero activations, while the quality of the description is proportional to the reconstruction loss.
The $\varphi$ metric is also related to the rate-distortion theory~\cite{burger1971rate}, in which the maximum distortion is defined according to human perception, which however inevitably introduces a bias.
There is also relation with the field of Compressed Sensing~\cite{donoho2006compressed} in which the sparsity of the data is exploited allowing us to reconstruct it with fewer samples than the Nyquist-Shannon theorem requires and the field of Robust Feature Extraction~\cite{kim2013deep} where robust features are generated with the aim to characterize the data.
Olshausen et al.~\cite{olshausen1996emergence} presented an objective function that considers subjective measures of sparseness of the activation maps, however in this work we use the direct measure of compression ratio.
Previous work by~\cite{zhang2017ecg} have used a weighted combination of the number of neurons, percentage root-mean-squared difference and a correlation coefficient for the optimization function of a FNN as a metric but without taking consideration the number of non-zero activations.

A limitation of SANs is the use of varying amplitude-only kernels, which are not sufficient for more complex data and also do not fully utilize the compressibility of the data.
A possible solution would be using a grid sampler~\cite{jaderberg2015spatial} on the kernel allowing it to learn more general transformations (such as scale) than simple amplitude variability.
However, additional kernel properties should be chosen in correspondence with the $\varphi$ metric i.e.\ the model should compress more with decreased reconstruction loss.

\section{Conclusions and future work}\label{sec:conclusions}
In this paper first we defined the $\varphi$ metric to evaluate how well do models trade-off reconstruction loss with compression.
We then defined SANs which have minimal structure and with the use of sparse activation functions learn to compress data without losing important information.
Using Physionet datasets and MNIST we demonstrated that SANs are able to create high quality representations with interpretable kernels.

The minimal structure of SANs makes them easy to use for feature extraction, clustering and time-series forecasting.
Other future work on SANs include:
\begin{itemize}
	\item Applying cosine annealing to the extrema distance in order to increase the degree of freedom of the kernels.
	\item Imposing the minimum extrema distance along all similarity matrices for multiple kernels, thus making kernels compete for territory.
	\item Applying dropout at the activations in order to correct weights that have overshot, especially when they are initialized with high values.
		However, the effect of dropout on SANs would generally be negative since SANs have much less weights than DNNs thus need less regularization.
	\item Using SANs with dynamically created kernels that might be able to learn multimodal data from variable sources (e.g.\ from ECG to respiratory) without destroying previous learned weights.
\end{itemize}

\section*{ACKNOWLEDGMENT}
This work was supported by the European Union's Horizon 2020 research and innovation programme under Grant agreement 769574.
We gratefully acknowledge the support of NVIDIA with the donation of the Titan X Pascal GPU used for this research.

\bibliographystyle{IEEEtran}
\bibliography{ms.bib}

\end{document}